\documentclass{article} 
\usepackage{utils/iclr2026_conference,times}

\usepackage{amsmath,amsfonts,bm}

\def\eqref#1{equation~\ref{#1}}

\def\1{\bm{1}}

\DeclareMathAlphabet{\mathsfit}{\encodingdefault}{\sfdefault}{m}{sl}
\SetMathAlphabet{\mathsfit}{bold}{\encodingdefault}{\sfdefault}{bx}{n}

\usepackage[utf8]{inputenc} 
\usepackage[T1]{fontenc}    
\usepackage{url}
\usepackage{booktabs}  
\usepackage{nicefrac}       
\usepackage{xcolor}         
\usepackage{soul}

\usepackage[ruled,vlined,linesnumbered]{algorithm2e}
\SetAlgoNlRelativeSize{0} 
\SetNlSty{}{}{}           
\DontPrintSemicolon       
\IncMargin{0.1em}           
\usepackage{tikz}

\usetikzlibrary{backgrounds, calc}
\pgfdeclarelayer{background}
\pgfsetlayers{background,main}
\usepackage{xspace} 
\usepackage{amssymb} 
\usepackage{multirow} 
\usepackage{subcaption}
\usepackage{algpseudocode}
\usepackage{wrapfig}
\usepackage{appendix}

\usepackage{hyperref}
\usepackage{graphicx}
\usepackage{amsmath}
\usepackage{amsthm}
\usepackage[export]{adjustbox} 
\definecolor{color1}{RGB}{62, 9, 136} 
\definecolor{color2}{RGB}{182, 55, 121} 
\definecolor{color3}{RGB}{239, 191, 92} 

\newcommand{\myparagraph}[1]{\noindent \textbf{#1}}

\newcommand{\vct}[1]{\ensuremath{\boldsymbol{#1}}}

\newcommand{\set}[1]{\ensuremath{\mathcal{#1}}}

\newcommand{\ie}{{i.e.}\xspace}

\newcommand{\ellzero}{$\ell_0$\xspace}

\newcommand{\ellinf}{$\ell_{\infty}$\xspace}
\newcommand{\robustloss}{$\hat{\mathcal{L}}$\xspace}

\newcommand{\ssap}{S2AP\xspace}

\title{\ssap: Score-space Sharpness Minimization for Adversarial Pruning}

\author{%
\makebox[\textwidth][c]{%
\begin{tabular}{c}\\[1em]
\textbf{Giorgio Piras}$^{1}$ \quad
\textbf{Qi Zhao}$^{2}$ \quad
\textbf{Fabio Brau}$^{1}$ \quad
\textbf{Maura Pintor}$^{1}$ \\
\textbf{Christian Wressnegger}$^{2}$ \quad
\textbf{Battista Biggio}$^{1}$
\end{tabular}}\\[3em]
\makebox[\textwidth][c]{\small
$^{1}$University of Cagliari \quad
$^{2}$KASTEL Security Research Labs, Karlsruhe Institute of Technology}
}

\iclrfinalcopy 
\begin{document}

\maketitle
\begin{abstract}
Adversarial pruning methods have emerged as a powerful tool for compressing neural networks while preserving robustness against adversarial attacks. 
These methods typically follow a three-step pipeline: (i) pretrain a robust model, (ii) select a binary mask for weight pruning, and (iii) finetune the pruned model.
To select the binary mask, these methods minimize a robust loss by assigning an importance score to each weight, and then keep the weights with the highest scores.
However, this score-space optimization can lead to sharp local minima in the robust loss landscape and, in turn, to an unstable mask selection, reducing the robustness of adversarial pruning methods. 
To overcome this issue, we propose a novel plug-in method for adversarial pruning, termed Score-space Sharpness-aware Adversarial Pruning (\ssap).
Through our method, we introduce the concept of score-space sharpness minimization, which operates during the mask search by perturbing importance scores and minimizing the corresponding robust loss. 
Extensive experiments across various datasets, models, and sparsity levels demonstrate that \ssap effectively minimizes sharpness in score space, stabilizing the mask selection, and ultimately improving the robustness of adversarial pruning methods.
\end{abstract}

\section{Introduction}\label{introduction}
Deep neural networks are susceptible to adversarial attacks, which entail optimizing an input perturbation added to the original sample to induce a misclassification~\citep{biggio13-ecml, szegedy14-iclr}. 
Besides robustness against adversarial examples, networks are often required to be compact and suitable for resource-constrained scenarios~\citep{liu2023lessons}, where the model's dimension cannot be chosen at hand but requires respecting a given constraint. 
In this regard, neural network pruning~\citep{lecun89_nips} represents a powerful compression method by removing redundant or less impactful parameters according to a desired sparsity rate and, as a result, allowing the preservation of much of the performance of a dense model counterpart~\citep{blalock20_mlsys}. 

Adversarial Pruning (AP) methods aim to fulfill this twofold requirement, thus extending model compression to the adversarial case, by removing parameters less responsible for adversarial robustness drops~\citep{piras2024_ap}. 
Following the three-step pipeline from~\cite{han15_nips}, AP methods often operate by pruning a pretrained robust model and then finetuning the resulting sparse model via adversarial training~\cite{sehwag20_nips, zhao23_iclr, madry18-iclr}. 
While prior work extended na\"ive pruning heuristics to robustness, such as based on the lowest weight magnitude (LWM) of robust models~\citep{han15_nips, sehwag19_towards}, recent approaches proposed different strategies to quantify each parameter's importance, and thus select an optimized pruning mask accordingly.
These methods, such as HARP~\citep{zhao23_iclr} and HYDRA~\citep{sehwag20_nips}, use real-valued importance scores, one for each model's weight, indicating how much robust loss degrades based on that parameter's removal.  
These scores are then optimized during the pruning stage by: (i) computing the robust loss using the top-$k$ parameters in the forward pass (where $k$ is the desired sparsity); and (ii) updating each parameter's importance in the backward pass. 
This procedure circumvents the intractability of the binary mask optimization problem imposed by the \ellzero constraint on the weights (\ie, the desired sparsity). Hence, it enables a parameter selection process based on the scores minimizing a robust objective, yielding a final mask with enhanced adversarial robustness.

Consequently, however, the effectiveness of the pruning mask in preserving robustness is strongly dependent on importance-score optimization, where minor score variations can lead to large changes in the selected top-$k$ parameters due to the combinatorial nature of the mask selection.
This volatility, potentially altering the mask selection and undermining the model's robustness, highlights the need for a smoother, more stable optimization landscape.
In this regard, recent advances in neural network training suggest that explicitly minimizing sharpness in the loss landscape can foster not only generalization~\citep{foret21_iclr}, but also adversarial robustness~\citep{wu20_nips, stutz21_iccv}. 
These approaches, such as Adversarial Weight Perturbations (AWP)~\citep{wu20_nips} for adversarial robustness, work by perturbing the network parameters (\ie, the weights) and minimizing the corresponding loss to reduce sharpness and improve performance.

\begin{figure}[t]
  \centering
  \begin{subfigure}[t]{0.31\linewidth}
    \centering
    \includegraphics[width=\linewidth]{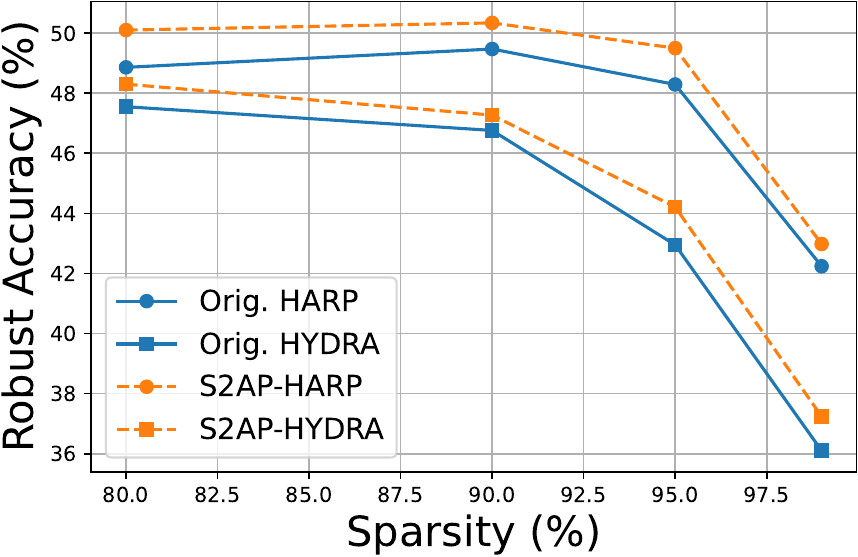}
    \caption{}
    \label{fig:intro_robustness}
  \end{subfigure}%
  \hspace{0.01\linewidth}
  \begin{subfigure}[t]{0.31\linewidth}
    \centering
    \includegraphics[width=\linewidth]
      {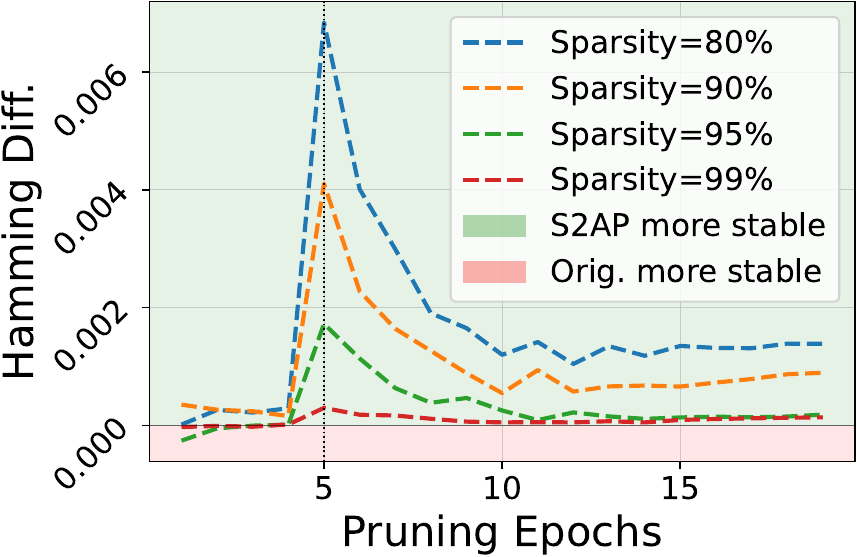}
    \caption{}
    \label{fig:intro_hamming}
  \end{subfigure}%
  \hspace{0.01\linewidth}
  \begin{subfigure}[t]{0.32\linewidth}
    \centering
    \includegraphics[width=\linewidth]
      {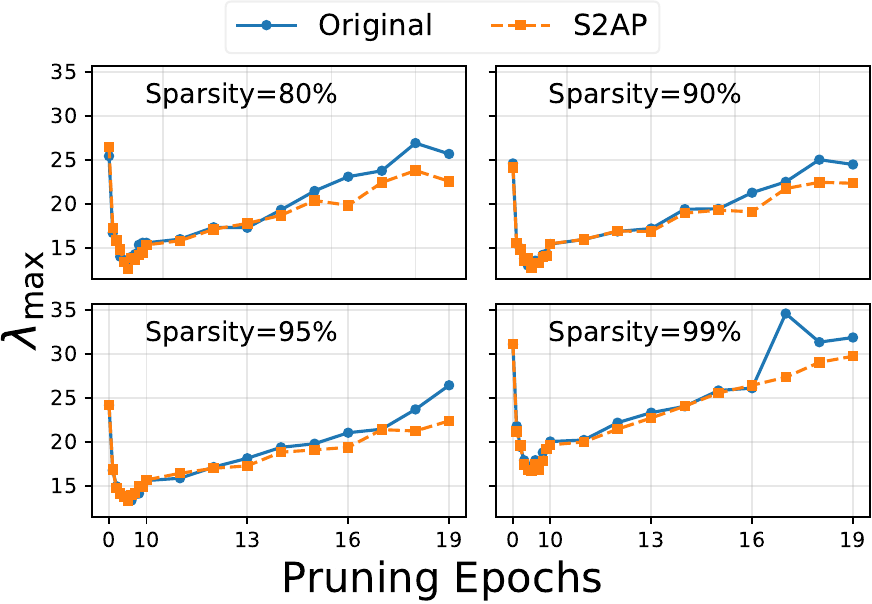}
    \caption{}
    \label{fig:intro_eigenvalues}
  \end{subfigure}

  \caption{Effects of \ssap on a ResNet18 CIFAR10 model. (a) Improved robust accuracy of pruned models. (b) Enhanced mask stability (quantified as Hamming distance $h$, i.e., measuring how much each mask $\vct m_t$ across pruning epochs changes compared to the first computed mask $\vct m_0$). We subtract and plot $h_{orig}-h_{\ssap}$, thus yielding positive values where \ssap is more stable (green area), and negative values vice versa (red area). \ssap enhances mask stability, particularly after pruning epoch 5 when warm-up ends and explicit sharpness minimization begins. (c) Minimized sharpness in the robust loss landscape (where the largest eigenvalue $\lambda_{max}$ indicates more sharpness). 
  }
  \label{fig:intro_figure}
\end{figure}

Inspired by these findings, we extend the concept of sharpness minimization in adversarial robustness beyond the traditional parameter-space setting, where weights are perturbed, to the novel context of importance score optimization. 
We thereby propose a \textit{score-space} sharpness minimization approach for adversarial pruning methods, that we define as Score-space Sharpness-aware Adversarial Pruning (\ssap), which reduces the sharpness of the loss landscape parameterized by importance scores, stabilizing the mask selection and improving adversarial robustness of pruned models.
Crucially, \ssap is implemented as a plug-in, allowing seamless integration into existing AP methods (and any other score-based approach) without altering their core logic or loss formulation. Overall, our main contributions are organized as follows: 
\begin{itemize}
    \item[(i)] we present the \ssap method (\autoref{sect:SSAP}), discussing its algorithm in a step-by-step approach; 
    \item[(ii)] we then demonstrate, across multiple architectures, datasets, and sparsity rates, how \ssap improves robustness of adversarial pruning methods (\autoref{sect:exp_robustness});
    \item[(iii)] we finally show, on the same comprehensive setup, the minimized sharpness in the score-space landscape and the induced mask search stability (\autoref{sect:exp_sharpness}).
\end{itemize}
A preview of the discussed effects and results can be seen in~\autoref{fig:intro_figure}, where we show the improved robustness of \ssap (\autoref{fig:intro_robustness}), the stabilized mask selection based on the masks' Hamming distances (\autoref{fig:intro_hamming}), and the minimized sharpness based on the largest eigenvalue (\autoref{fig:intro_eigenvalues}).

\section{Adversarial Pruning and Score-Space}\label{sect:background}

\myparagraph{Notation.} Let us denote with $\set D = \left\{(\vct x_i, y_i)\right\}_{i=1}^n$ a training set of $n$ $d$-dimensional samples $\vct x \in \set X = [0,1]^d$ along with their labels $y \in \set Y = \{1,\dots,C\}$. For a network $f(\cdot\,; \vct w)$ with parameters $\vct w \in \mathbb{R}^p$, we define the average loss function computed on the dataset $\set D$ (or on a batch) as $\set L(\vct w, \set D) = 1/n \sum_i \ell(y_i, f(\vct x_i; \vct w))$, being $\ell$ any suitable sample-wise loss and $f$ the $C$ logits of the network. 

\myparagraph{Adversarial Robustness.} 
Machine Learning (ML) models are susceptible to adversarial attacks~\citep{biggio13-ecml,szegedy14-iclr}, which create input samples misclassified by the attacked model. 
In this regard, adversarial training is considered the go-to defense, minimizing a given robust loss $\hat{\mathcal{L}}$ defined as the inner maximization in the following optimization problem:
\begin{equation}\label{eq:at}
\min_{\vct w} \:  \hat{\mathcal{L}}(\vct w, \set D), \: \quad  \hat{\mathcal{L}} (\vct w, \set D) = \frac{1}{n} \sum_{i=1}^n \max_{\|\vct \delta_i\| \leq \epsilon} \ell(y_i, f(\vct x_i + \vct \delta_i; \vct w)),
\end{equation}

where $\vct x_i + \vct \delta_i \in [0,1]^d, \forall i$, i.e., that each perturbed sample still lies in the sample space upon adding an adversarial perturbation $\vct \delta$ bounded by a given $\ell_p$ bound $\epsilon$.  

\myparagraph{Adversarial Pruning Methods.} Pruning aims to reduce the size of a network by removing its parameters (e.g., weights) while preserving performance~\citep{lecun89_nips}. 
Similarly, Adversarial Pruning (AP) methods aim to reduce model size while preserving robustness against adversarial attacks~\citep{piras2024_ap}. Recent AP methods proposed solving the following optimization problem: 
\begin{align}
\label{eq:ap_constrained}  \underset{\|\vct m\|_0 \leq k}{\min} \:  \hat{\mathcal{L}}(\vct w \odot \vct m, \set D), 
\end{align}
where $\vct m \in \{0,1\}^p$ is a $p$-dimensional mask constrained to have $k$ non-zero entries. The mask is element-wise multiplied by the weights $\vct w$, ensuring that the pruned model satisfies the sparsity rate $k$. 
However, the sparsity constraint makes~\autoref{eq:ap_constrained} a non-convex, combinatorial problem. AP methods like HARP~\citep{zhao23_iclr}, HYDRA~\citep{sehwag20_nips}, thus solve it by relaxing the sparsity constraint through the use of \textit{importance scores}.

\myparagraph{Importance Scores.}
During the pruning stage, while weights are kept invariant, optimizing importance scores amounts to defining a vector of continuous values $\vct s \in \mathbb{R}^p$, initialized proportionally to the weights, which are learnable and optimized with respect to the robust loss \robustloss as follows: 
\begin{align}
\label{eq:ap_scores}\min_{\vct s} \:  \hat{\mathcal{L}}(\vct w \odot M(\vct s, k), \set D), 
\end{align}
where \robustloss is computed, given $k$, through a masking function $M(\vct s,k)$ that assigns $1$ only to the top-k entries of $\vct{s}$, thus imposing sparsity.
Let us remark that such an optimization procedure is non-trivial: in the forward pass, the loss is computed using the top-$k$ parameters as $\hat{\mathcal{L}}(\vct w \odot M(\vct s, k), \set D)$; during backpropagation, these methods adopt a straight-through estimator (STE) substituting $\partial M/\partial \vct s$ with $1$ (\ie, the identity) following~\cite{ramanujan20_ste}. This method enables propagating the gradient through the non-differentiable mask and optimizing each score according to its importance. We thus define as \textbf{score‑space} the $p$‑dimensional space $\mathbb R^{p}$ spanned by the importance scores $\vct s$, and study the robust loss landscape $\hat{\mathcal{L}}(\vct w\odot M(\vct s,k),\mathcal D)$ defined over it.

\myparagraph{Formulation Generality.}
The formulation of~\autoref{eq:ap_scores} encompasses all AP methods based on importance-score optimization. 
Different methods can, however, define different loss functions (that we generalize through \robustloss). 
This is the case of HARP~\citep{zhao23_iclr}, which defines additional penalty terms allowing for optimizing layer-wise sparsity.  
We specify that our formulation unifies different loss formulations from various AP methods, and as we will describe in the next section, the proposed \ssap can ``wrap'' any AP method based on importance-score optimization.

\section{\ssap: Minimizing Score-Space Sharpness}\label{sect:SSAP} 

From~\autoref{sect:background}, it becomes evident that score optimization on a robust loss is the core logic of adversarial pruning.  
We improve such an approach by minimizing score-space sharpness. Hence, our Score-space Sharpness-aware Adversarial Pruning (\ssap) method avoids that small score shifts induce relevant mask changes, thus stabilizing the pruning process and increasing adversarial robustness.
Following~\autoref{eq:ap_scores}, and given the sharpness minimization approach from~\cite{wu20_nips}, we define the \ssap problem as follows: 
\begin{align}
\label{eq:SSAP}\vct s^* \in \underset{\vct s}{\arg \min} \: \underset{\vct z}{\max} \: \hat{\mathcal{L}}(\vct w \odot M(\vct s + \vct z, k), \mathcal{D}) \, ,\\
\label{eq:gamma_bound} \text{where } \|\vct z_l\| \leq \gamma \|\vct s_l\|, 
\end{align}
and $\gamma$ constraints the \textit{score perturbation} $\vct z$ applied on $\vct s$, scaling it w.r.t. the norm of the scores of each layer $l$. \ssap solves such optimization through \autoref{alg:s2ap}, as detailed below. Note that the sections of~\autoref{alg:s2ap} outside the orange box are common to AP methods (cf.~\autoref{sect:background}). 

\begin{algorithm}[t]
 \SetKwInOut{Input}{Input}
 \SetKwInOut{Output}{Output}
 \SetKwComment{Comment}{$\triangleright$\ }{}
 \DontPrintSemicolon
 \caption{Score-Sharpness-aware Adversarial Pruning.}
 \label{alg:s2ap}
 \Input{$\vct w \in \mathbb{R}^p$, initial weights; $\vct s \in \mathbb{R}^p$, set of importance scores;  $M(\vct s, k)$, masking function with pruning rate $k$; $\vct x$, training inputs samples; $\eta$, learning rate; $I$, number of iterations; $L$, number of layers; $\gamma$, perturbation scaling factor; $\hat{\mathcal{L}}$, robust loss.} 
 \Output{Binary mask $\vct m^* \in \{0,1\}^d$.}

 Initialize parameters $\vct s = \texttt{scale}(\vct w), \; \vct x'_i \gets \vct x, \;  \vct s^* \gets \vct s, \; \vct z \gets 0$\; \label{line:init}
 \For{$i\gets1$ \KwTo $I$}{
   Generate adversarial examples on pruned model $\vct x_i' \;\gets\; \vct x_i + \vct \delta_i$\; \label{line:adv_ex}
   Compute robust loss on pruned model $\hat{\mathcal{L}}(\vct s)=\hat{\mathcal{L}}(\vct w \odot M(\vct s,k), \mathcal{D})$\; \label{line:robust_loss}

    \tikz[remember picture] \node[coordinate] (ssapstart) {};

    Generate score-space perturbation $\vct z \;\gets\ \vct z + \eta (\nabla_{z}\hat{\mathcal{L}}(\vct s+\vct z)/\|\nabla_{z}\hat{\mathcal{L}}(\vct s+\vct z)\|)$\; \label{line:z_update}
    \For{$l \gets 1$ \KwTo $L$} {
     \If{$\|\vct z^{(l)}\| > \gamma \,\|\vct s^{(l)}\|$} {
       Project perturbation $\vct z^{(l)} \;\gets\; \bigl(\gamma\,\|\vct s^{(l)}\|\;/\;\|\vct z^{(l)}\|\bigr)\,\vct z^{(l)}$\;\label{line:layer_proj}
     }
    }
    Update scores $\vct s \;\gets\ \vct s - \eta(\nabla_{s}\hat{\mathcal{L}}(\vct s+\vct z)/\|\nabla_{s}\hat{\mathcal{L}}(\vct s+\vct z)\|)$\; \label{line:s_update}
    Restore scores $\vct s \;\gets\ \vct s - \vct z$\; \label{line:z_undo}

    \tikz[remember picture] \node[coordinate] (ssapend) {};

   \If{$\hat{\mathcal{L}}(\vct s)<\hat{\mathcal{L}}(\vct s^*)$} {
     Update best loss $\hat{\mathcal{L}}(\vct s^*) \gets \hat{\mathcal{L}}(\vct s)$\; \label{line:update_best}
   }
 }
 \textbf{return} $\vct m^* \gets M(\vct s^*, k)$\; \label{line:return_best}
\end{algorithm}
\begin{tikzpicture}[remember picture, overlay]
  \begin{pgfonlayer}{background}
    \coordinate (ssapnw) at ($(ssapstart) + (38.1em, 0.5ex)$);
    \coordinate (ssapse) at ($(ssapend) + (-0.2em, 0.5ex)$);
    \path[draw=orange!89!black, rounded corners=3pt, line width=1pt]
      (ssapnw) rectangle (ssapse);
  \end{pgfonlayer}
  
  \node[anchor=north east, font=\normalsize\sffamily\bfseries, text=black] 
    at ($(ssapse) + (38.2em, 03.5ex)$) {\ssap};
\end{tikzpicture}

\myparagraph{Generating Adversarial Examples.} 
We initialize, in~\autoref{line:init}, the set of importance scores $\vct s$ proportionally to $\vct w$ through \texttt{scale}, which scales the scores proportionally to the weights' magnitude.  
This enables creating a pruned model ($f(\vct w \odot M(\vct s, k)$) through which we compute the adversarial examples $\vct x'$ (\autoref{line:adv_ex}) 
using the \ellinf PGD attack~\citep{madry18-iclr}. Following~\autoref{eq:at}, we thus craft a perturbation $\vct \delta$ constrained on $\epsilon$. 
Computing the adversarial examples allows defining a robust loss \robustloss which we denote, for brevity and emphasis on the scores, as $\hat{\mathcal{L}}(\vct s)$ in~\autoref{line:robust_loss}.

\myparagraph{Score-Space Perturbation.} 
Defining a robust loss and creating adversarial examples is a common step of score-based AP methods. 
During the pruning stage, in fact, these methods' weights are left unchanged while importance scores $\vct s$ are optimized according to a robust objective to find the best mask $\vct m = M(\vct s, k)$. 
Through \ssap, we are interested in minimizing the score-space sharpness.
Hence, before the standard score optimization, when using \ssap we craft a score-space perturbation (\autoref{line:z_update}) in one single iteration, aiming to shift the loss in score space from the $i$-th iteration's local minima towards a point of higher loss.
We thus create a \textit{worst-case} score perturbation.

In~\autoref{line:layer_proj}, we iterate over the $L$ layers of the network and project our perturbation $\vct z$ in a bound defined by $\gamma$.
More precisely, according to the layer's score magnitude $\|\vct s^{(l)}\|$, we scale $\vct z^{(l)}$ to $(\gamma\,\|\vct s^{(l)}\|\;/\;\|\vct z^{(l)}\|)\,\vct z^{(l)}$ if $\|\vct z^{(l)}\| > \gamma \,\|\vct s^{(l)}\|$, which corresponds to projecting back the perturbation into the ``ball'' defined by $\gamma$ when exceeding, and leave as is otherwise.
The layer-wise projection primarily addresses the numeric differences across layers. 
Without per-layer scaling, the magnitude of the generated perturbation $\vct z$ can be perceived differently across layers, leading to either no effect or numerical overflow. 
A layer-wise projection instead keeps every layer's perturbation proportional to its current score norm, preserving well-conditioned updates and preventing disparity across layers.  

\myparagraph{Score Update.} 
Once the score perturbation $\vct z$ is computed, we evaluate the gradient at the perturbed scores $\vct s + \vct z$ (\autoref{line:s_update}), and take an optimization step to move $\vct s$ in the direction that, in turn, reduces sharpness. 
After optimizing \robustloss, \ssap ends by removing the previously applied perturbation to restore the original reference point $\vct s$ for the next iteration (\autoref{line:z_undo}). 
We specify that also the score update of~\autoref{line:s_update} is common to AP methods. 
However, instead of updating scores based on the loss computed on score space $\hat{\mathcal{L}}(\vct s)$, \ssap enables a ``sharpness-aware'' update on perturbed score space $\hat{\mathcal{L}}(\vct s + \vct z)$. 
Finally, through~\autoref{line:update_best} and~\autoref{line:return_best}, we save $\vct s^*$ corresponding to the lowest \robustloss and return the best mask $\vct m^*$ via the function $M(\vct s^*, k)$, which is finally multiplied to the pretrained weights.

\myparagraph{\ssap Finetuning.} 
After defining mask $\vct m^*$ and pruning the model, some of the AP methods we enhance with \ssap finetune the pruned weights to restore performance using a robust objective~\citep{han15_nips}. 
In \ssap, we choose to finetune the pruned model by aligning the objective with the score-space sharpness minimization implemented while pruning. 
Hence, we choose to minimize sharpness using the AWP~\citep{wu20_nips} approach applied on the classical weight-space: 
\begin{align}
\label{eq:SSAP_finetuning}\vct w^* \in \underset{\vct w}{\arg \min} \: \underset{\vct \nu}{\max} \:\hat{\mathcal{L}}((\vct w+\vct \nu) \odot \vct m^*) \, ,\\
\label{eq:gamma_bound2} \text{where } \|\vct \nu_l\| \leq \gamma \|\vct w_l\|,
\end{align} 
and $\nu$, in this case, is a weight perturbation added to the preserved weights according to $\vct m^*$ found through~\autoref{alg:s2ap}. 
Therefore, instead of perturbing all the weights as in typical sharpness minimization, we add a perturbation only to the top-$k$ weights according to the mask found in the previous step, and project the perturbation based on the layers' weight magnitude.
We provide a more detailed explanation of our finetuning approach in~\autoref{app:apdx_finetuning}.

\section{Experiments}\label{sect:experiments} 
\ssap minimizes score-space sharpness, building upon the observation that a smoother loss landscape enhances adversarial robustness. 
In turn, after describing the general experimental setup (\autoref{sect:exp_setup}), we show and discuss the robustness of \ssap on adversarial pruning methods (\autoref{sect:exp_robustness}), and then analyze the effect of \ssap on score-space sharpness minimization and mask selection stability (\autoref{sect:exp_sharpness}). More experiments can be found in~\autoref{app:appendix_ssap}, \autoref{app:appendix_additional_exps}, and~\autoref{app:appendix_sharpness}.
 
\subsection{Experimental Setup}\label{sect:exp_setup}

\myparagraph{AP Methods, Models, and Datasets.}
We test \ssap on the HARP, HYDRA, and Robust-Lottery Ticket Hypothesis (RLTH) adversarial pruning methods~\citep{zhao23_iclr, sehwag20_nips, fu2021drawing}, while comparing to the original implementations (Orig.). 
These approaches are all based on the optimization of importance scores summarized in~\autoref{eq:ap_scores}.
However, while HARP and HYDRA start from a robust pretrained model, and, after pruning, finetune the pruned model, RLTH tests the LTH on a randomly initialized model and does not finetune the resulting pruned parameterization. We show RLTH results in~\autoref{app:appendix_additional_exps}
We choose $80\%, 90\%, 95\%$, and $99\%$ as sparsity rates, indicating the rate of pruned parameters. We employ the ResNet18~\citep{he2016deep}, VGG16~\citep{simonyan2015very}, and WideResNet-28-4~\citep{zagoruyko2016wide} architectures on both the CIFAR10~\citep{krizhevsky2009learning} and SVHN~\citep{netzer2011reading} datasets. In addition, we test HARP and HYDRA on the larger-scale ImageNet~\citep{deng2009imagenet} dataset using the ResNet50 architecture (we refrain from testing RLTH on ImageNet, as with no finetuning, the accuracy is too low with moderate epochs). Finally, we prune a vision transformer (ViT) with a patch size of $4\times4$, resulting in $64$ tokens for $32\times32$ images, to $20\%$, $40\%$, and $60\%$ sparsity. It comprises $8$ transformer layers, $6$ attention heads per layer, and a hidden dimensionality of $384$. The MLP blocks have an expansion ratio of $4$, with a hidden dimension of $1536$.

\myparagraph{Adversarial Training and Evaluation.} 
We pretrain, prune, and finetune the models with HARP and HYDRA (prune only for RLTH) using the TRADES loss~\citep{zhang19_icml} (pretrained models' results are shown in~\autoref{app:apdx_pretraining}).
During adversarial training, we generate adversarial examples using \ellinf PGD-10 with perturbation size $\epsilon=8/255$ and step-size $\alpha=2/255$.
Similarly, we evaluate robustness using the AutoAttack (AA~\citep{croce20_autoattack}) ensemble with \ellinf perturbation bound $\epsilon=8/255$ for every adversarial robustness evaluation.
For HARP and HYDRA, we pretrain and finetune in $100$ epochs, while we prune for $20$ epochs. Also, we search for the RLTH tickets in $20$ epochs. Of these $20$ epochs, for each method, \ssap is applied after $5$ warm-up epochs. For completeness, we discuss the computational cost of pruning with \ssap in~\autoref{app:apdx_overhead}.

\myparagraph{\ssap Setup.}
We use the same adversarial training setup as the original methods to prune with \ssap. Also, we find one step to be sufficient for finding a score perturbation, as in~\cite{wu20_nips}. However, we must specify a $\gamma$ value to design the layer-wise perturbation projection. For ResNet18 and WideResNet on CIFAR10, we set $\gamma=0.001$; for VGG16 on CIFAR10 and SVHN, $\gamma=0.0025$; for ResNet18 on SVHN, $\gamma=0.0075$; for WideResNet on SVHN, $\gamma=0.005$; and finally, for ResNet50 on ImageNet, we set $\gamma=0.0075$. The same $\gamma$ is used to bound weight perturbation for \ssap finetuning in HARP and HYDRA. For ViTs, we choose gamma $0.0025$. 
We select the $\gamma$ value according to the highest robust accuracy, and discuss its selection in~\autoref{app:apdx_gamma}.

\begin{table*}[ht]
\centering
\caption{CIFAR-10 results. We show the clean/robust$_{\pm{std}}$ accuracy and the difference between Orig. and \ssap robust generalization gap ($\Delta$). In bold, the model with the highest robustness.}
\label{tab:s2ap_cifar10_robust_clean}
\resizebox{\textwidth}{!}{%
\begin{tabular}{@{}lccccccc@{}}
\toprule
\multirow{2}{*}{Network} &
\multirow{2}{*}{Sparsity} & 
\multicolumn{3}{c}{HARP} & 
\multicolumn{3}{c}{HYDRA} \\
\cmidrule(lr){3-5} \cmidrule(l){6-8}
& & Orig. & \ssap & Gap $\Delta$ & Orig. & \ssap & Gap $\Delta$ \\
\midrule
\multirow{4}{*}{ResNet18}
& 80\% & 81.26 / 48.86$_{\pm0.16}$ & \textbf{81.36 / 50.10$_{\pm0.21}$} & +1.14 & 80.73 / 47.55$_{\pm0.81}$ & \textbf{81.47 / 48.30$_{\pm0.91}$} & +0.01 \\
& 90\% & 81.62 / 49.47$_{\pm0.24}$ & \textbf{82.10 / 50.34$_{\pm0.33}$} & +0.39 & 80.85 / 46.76$_{\pm1.34}$ & \textbf{80.89 / 47.27$_{\pm1.09}$} & +0.47 \\
& 95\% & 82.88 / 48.29$_{\pm0.44}$ & \textbf{82.68 / 49.50$_{\pm0.46}$} & +1.41 & 80.83 / 42.95$_{\pm1.38}$ & \textbf{80.14 / 44.21$_{\pm0.92}$} & +1.95 \\
& 99\% & 80.72 / 42.24$_{\pm0.13}$ & \textbf{81.46 / 42.98$_{\pm0.39}$} & +0.00 & 80.51 / 36.10$_{\pm1.41}$ & \textbf{80.93 / 37.24$_{\pm1.20}$} & +0.72 \\
\midrule
\multirow{4}{*}{VGG16} 
& 80\% & 78.49 / 45.20$_{\pm0.69}$ & \textbf{79.19 / 45.93$_{\pm0.34}$} & +0.03 & 77.10 / 44.63$_{\pm0.09}$ & \textbf{78.70 / 44.95$_{\pm0.12}$} & -1.28 \\
& 90\% & 80.54 / 45.53$_{\pm0.47}$ & \textbf{78.64 / 46.26$_{\pm0.41}$} & +2.63 & 77.65 / 43.07$_{\pm0.23}$ & \textbf{77.07 / 43.57$_{\pm0.06}$} & +1.08 \\
& 95\% & 78.70 / 44.74$_{\pm0.23}$ & \textbf{79.12 / 45.67$_{\pm0.11}$} & +0.51 & 76.79 / 40.75$_{\pm0.72}$ & \textbf{76.55 / 41.48$_{\pm0.83}$} & +0.97 \\
& 99\% & 77.85 / 41.38$_{\pm0.88}$ & \textbf{78.61 / 42.04$_{\pm0.36}$} & -0.10 & 75.10 / 33.24$_{\pm1.44}$ & \textbf{76.43 / 34.09$_{\pm1.04}$} & -0.48 \\
\midrule
\multirow{4}{*}{WRN28-4} 
& 80\% & 81.69 / 50.08$_{\pm0.67}$ & \textbf{81.73 / 51.28$_{\pm0.74}$} & +1.16 & 81.94 / 50.17$_{\pm0.68}$ & \textbf{82.37 / 50.79$_{\pm0.47}$} & +0.19 \\
& 90\% & 82.02 / 50.52$_{\pm0.51}$ & \textbf{82.31 / 51.83$_{\pm0.71}$} & +1.02 & 81.24 / 50.17$_{\pm0.35}$ & \textbf{82.29 / 50.40$_{\pm0.67}$} & -0.82 \\
& 95\% & 82.47 / 50.57$_{\pm0.76}$ & \textbf{82.49 / 51.04$_{\pm0.58}$} & +0.45 & 81.42 / 49.22$_{\pm0.21}$ & \textbf{81.90 / 49.40$_{\pm0.78}$} & -0.30 \\
& 99\% & 76.14 / 44.68$_{\pm0.82}$ & \textbf{76.29 / 44.93$_{\pm0.27}$} & +0.10 & \textbf{74.66 / 42.28$_{\pm0.58}$} & 74.00 / 42.01$_{\pm0.64}$ & +0.39 \\
\bottomrule
\end{tabular}%
}
\end{table*}

\begin{table*}[ht]
\centering
\caption{SVHN results. We show the clean/robust$_{\pm{std}}$ accuracy and the difference between Orig. and \ssap robust generalization gap ($\Delta$). In bold, the model with the highest robustness.}
\label{tab:s2ap_svhn_robust_clean}
\resizebox{\textwidth}{!}{%
\begin{tabular}{@{}lccccccc@{}}
\toprule
\multirow{2}{*}{Network} &
\multirow{2}{*}{Sparsity} & 
\multicolumn{3}{c}{HARP} & 
\multicolumn{3}{c}{HYDRA} \\
\cmidrule(lr){3-5} \cmidrule(l){6-8}
& & Orig. & \ssap & Gap $\Delta$ & Orig. & \ssap & Gap $\Delta$ \\
\midrule
\multirow{4}{*}{ResNet18} 
& 80\% & 92.55 / 40.06$_{\pm1.03}$ & \textbf{91.53 / 41.50$_{\pm1.05}$} & +2.46 & 92.71 / 42.56$_{\pm1.02}$ & \textbf{92.69}/ \textbf{43.72}$_{\pm1.07}$ & +1.18 \\
& 90\% & 91.61 / 40.14$_{\pm0.82}$ & \textbf{91.07 / 41.33$_{\pm0.26}$ }& +1.73 & \textbf{91.90} / \textbf{41.83}$_{\pm0.65}$ & 91.63 / 41.58$_{\pm0.30}$ & +0.02 \\
& 95\% & 87.53 / 38.16$_{\pm0.66}$ & \textbf{88.68} / \textbf{38.75}$_{\pm0.19}$ & -0.56 & 90.33 / 40.53$_{\pm0.16}$ & \textbf{90.63} / \textbf{40.86}$_{\pm0.28}$ & +0.03 \\
& 99\% & 88.42 / 35.24$_{\pm0.57}$ & \textbf{89.71} / \textbf{36.12}$_{\pm0.76}$ & -0.41 & 87.89 / 40.83$_{\pm0.83}$ & \textbf{88.63} / \textbf{41.10}$_{\pm0.28}$ & -0.47 \\
\midrule
\multirow{4}{*}{VGG16} 
& 80\% & 86.36 / 47.28$_{\pm1.11}$ & \textbf{87.80} / \textbf{49.69}$_{\pm1.05}$ & +0.97 & 85.75 / 46.13$_{\pm1.19}$ & \textbf{87.64} / \textbf{48.95}$_{\pm1.16}$ & +0.93 \\
& 90\% & 87.58 / 49.16$_{\pm1.12}$ & \textbf{87.77} / \textbf{49.49}$_{\pm1.19}$ & +0.14 & 86.22 / 48.04$_{\pm0.81}$ & \textbf{87.09} / \textbf{48.12}$_{\pm0.22}$ & -0.79 \\
& 95\% & 86.95 / 49.16$_{\pm0.29}$ & \textbf{86.98} / \textbf{49.28}$_{\pm0.58}$ & +0.09 & 86.10 / 45.95$_{\pm0.83}$ & \textbf{85.03} / \textbf{47.12}$_{\pm0.63}$ & +2.24 \\
& 99\% & 84.93 / 46.33$_{\pm0.51}$ & \textbf{84.73 / 46.61$_{\pm0.27}$} & +0.48 & \textbf{83.12} / \textbf{41.52}$_{\pm0.72}$ & 81.59 / 41.39$_{\pm0.46}$ & +1.40 \\
\midrule
\multirow{4}{*}{WRN28-4} 
& 80\% & 90.01 / 36.73$_{\pm1.02}$ & \textbf{90.65} / \textbf{43.53}$_{\pm0.61}$ & +6.16 & 95.24 / 42.95$_{\pm0.84}$ & \textbf{88.54} / \textbf{44.64}$_{\pm1.08}$ & +8.39 \\
& 90\% & \textbf{95.01} / \textbf{34.70}$_{\pm0.91}$ & 92.17 / 31.00$_{\pm0.76}$ & -0.86 & 93.35 / 36.29$_{\pm0.39}$ & \textbf{91.71} / \textbf{38.32}$_{\pm1.13}$ & +3.67 \\
& 95\% & 92.44 / 31.66$_{\pm0.77}$ & \textbf{94.46} / \textbf{33.15}$_{\pm0.72}$ & -0.53 & \textbf{89.55} / \textbf{43.99}$_{\pm0.65}$ & 90.43 / 38.89$_{\pm0.95}$ & -5.98 \\
& 99\% & 87.09 / 30.09$_{\pm0.83}$ & \textbf{88.47} / \textbf{36.26}$_{\pm1.12}$ & +4.79 & 93.05 / 31.24$_{\pm0.49}$ & \textbf{85.80} / \textbf{42.43}$_{\pm1.11}$ & +18.44 \\
\bottomrule
\end{tabular}%
}
\end{table*}

\subsection{Effect of \ssap on Adversarial Robustness}\label{sect:exp_robustness}
\ssap improves the robustness of adversarial pruning methods. 
We demonstrate such a result through \autoref{tab:s2ap_cifar10_robust_clean} for CIFAR10, \autoref{tab:s2ap_svhn_robust_clean} for SVHN, \autoref{tab:vit-cifar10} for transformers, and finally \autoref{tab:s2ap_imagenet_robust_clean} for ImageNet.
We further show results using channel pruning in~\autoref{app:apdx_structured}, and RLTH method in~\autoref{tab:s2ap_rlth_cifar10_svhn}

\begin{wraptable}{r}{0.45\textwidth}
\centering
\vspace{0pt} 
\caption{ViT on CIFAR-10 and HYDRA: clean / robust accuracy (\%) under different sparsity levels. Bold indicates the best between Orig. and S2AP.}
\label{tab:vit-cifar10}
\resizebox{0.45\textwidth}{!}{%
\begin{tabular}{llcc}
\toprule
Network & Sparsity (\%) & Orig. & S2AP \\
\midrule
\multirow{3}{*}{ViT}
& 20 & 63.93 / 26.45 & \textbf{64.53 / 27.85} \\
& 40 & 63.89 / 25.27 & \textbf{64.08 / 26.32} \\
& 50 & 63.02 / 24.71 & \textbf{63.87 / 25.86} \\
\bottomrule
\end{tabular}
}
\end{wraptable}

\myparagraph{Experimental Results.} 
In~\autoref{tab:s2ap_cifar10_robust_clean} for CIFAR10, across every sparsity level and method, \ssap consistently exceeds the robust accuracy of original methods. In general, across models, \ssap improves robustness up to 2 percentage points (p.p.). Importantly, these gains come with improved or negligible drops ($<0.3$ p.p.) in clean accuracy, as well as low error bars. To provide transparency on the trade-off between clean and robust performance, we also report the clean--robust generalization gap ($\Delta$), defined as the gap of Orig. minus that of \ssap. The gap measures the relative consistency between clean and robust accuracy, offering insight into how robust performance changes in relation to improvements or drops in clean accuracy. Across all settings, $\Delta$ remains mainly positive, showing that \ssap improves over Orig. without introducing a significant trade-off in generalization.
Overall, through our diverse experimental setup, we see the WideResNet model reaching higher robustness compared to the ResNet18 and VGG16 models, but still \ssap consistently outperforming competing methods.
A similar conclusion can be drawn for SVHN results in~\autoref{tab:s2ap_svhn_robust_clean} and ImageNet results on~\autoref{tab:s2ap_imagenet_robust_clean}. Again, \ssap consistently improves robustness across models, sparsities, and AP methods, with a comparable and often superior standard accuracy. We extend the \ssap evaluation to Vision Transformers in~\autoref{tab:vit-cifar10}. We remark how prior work on adversarial pruning has been limited to standard deep networks, thus marking this as a first experiment of AP methods on transformer-based architecture. We choose to prune with HYDRA, as the HARP method involves optimizing a layer-wise sparsity rate, which is not directly suited for transformer architectures and requires re-thinking the entire method. We prune all linear layers except for the final classification head, which is kept dense to ensure stable output mapping to class logits. We confirm the improved adversarial robustness on such kinds of architectures. Finally, we further validate the efficacy of \ssap by showing results for standard classification accuracy in~\autoref{app:apdx_clean}, and for robustness against common corruptions in~\autoref{app:apdx_corruptions}, thus validating \ssap in more general and external domains. 

\begin{wraptable}{r}{0.45\textwidth}
\centering
\vspace{-10pt} 
\caption{ImageNet results using ResNet50 across sparsity levels. Each cell shows clean/robust accuracy.}
\label{tab:s2ap_imagenet_robust_clean}
\resizebox{0.45\textwidth}{!}{%
\begin{tabular}{@{}lccc@{}}
\toprule
Network & Sparsity & Orig. & \ssap \\
\midrule
\multicolumn{4}{c}{\textbf{HARP}} \\
\midrule
\multirow{4}{*}{ResNet50} 
& 80\% & 61.48 / 33.01$_{\pm0.41}$ & \textbf{62.42 / 34.60$_{\pm0.82}$} \\
& 90\% & 54.93 / 24.05$_{\pm0.66}$ & \textbf{55.00 / 25.61$_{\pm0.57}$} \\
& 95\% & 40.74 / 21.12$_{\pm0.26}$ & \textbf{43.85 / 22.07$_{\pm0.26}$} \\
& 99\% & 28.65 / 12.92$_{\pm0.39}$ & \textbf{34.18 / 15.75$_{\pm0.76}$} \\
\midrule
\multicolumn{4}{c}{\textbf{HYDRA}} \\
\midrule
\multirow{4}{*}{ResNet50} 
& 80\% & 51.36 / 29.71$_{\pm0.48}$ & \textbf{56.16 / 31.11$_{\pm0.39}$} \\
& 90\% & 48.11 / 20.13$_{\pm0.33}$ & \textbf{54.92 / 24.23$_{\pm1.17}$} \\
& 95\% & 33.29 / 16.43$_{\pm0.67}$ & \textbf{34.19 / 17.93$_{\pm0.82}$} \\
& 99\% & 26.07 / 11.40$_{\pm0.20}$ & \textbf{27.47 / 12.67$_{\pm0.59}$} \\
\bottomrule
\end{tabular}%
}
\vspace{-10pt} 
\end{wraptable}

\myparagraph{Finetuning Ablation Study.}
In HARP and HYDRA, after selecting the mask through \ssap, we align the finetuning objective with the pruning one, thus finetuning by perturbing the weights via AWP~\citep{wu20_nips}, since scores are not used after pruning. 
We show in~\autoref{tab:all_bra} the "raw" mask adversarial robustness obtained before finetuning, thus the pruned model derived from multiplying the pretrained weights with the mask obtained in~\autoref{alg:s2ap}.
This comparison enables ablating the finetuning objective and verifying if the adversarial robustness improvement produced by \ssap is independent from finetuning. 
Our results highlight the higher robust accuracy of \ssap against the original AP methods throughout the different network and dataset combinations. 
In addition, we also discuss minimizing sharpness on the weights' loss landscape, and compare to \ssap, in~\autoref{app:apdx_awp_vs_ssap}.
\begin{table*}[ht]
\footnotesize
\centering
\caption{Mask robust accuracy (mean$_{\pm\text{std}}$) on CIFAR10 and SVHN across sparsity levels using ResNet18, VGG-16, and WideResNet-28-4.}
\label{tab:all_bra}
\resizebox{\textwidth}{!}{%
\begin{tabular}{@{}l c c c c c c c c c@{}}
\toprule
\multirow{2}{*}{Network} & \multirow{2}{*}{Sparsity} & \multicolumn{4}{c}{CIFAR10} & \multicolumn{4}{c}{SVHN} \\
\cmidrule(lr){3-6} \cmidrule(lr){7-10}
 & & \multicolumn{2}{c}{HARP} & \multicolumn{2}{c}{HYDRA} & \multicolumn{2}{c}{HARP} & \multicolumn{2}{c}{HYDRA} \\
\cmidrule(lr){3-4} \cmidrule(lr){5-6} \cmidrule(lr){7-8} \cmidrule(lr){9-10}
 & & Orig. & \ssap & Orig. & \ssap & Orig. & \ssap & Orig. & \ssap \\
\midrule
\multirow{4}{*}{ResNet18}
 & 80\% & 48.88$_{\pm0.73}$ & \textbf{49.55$_{\pm0.69}$} & 48.56$_{\pm0.66}$ & \textbf{48.98$_{\pm0.75}$} & 46.56$_{\pm0.66}$ & \textbf{49.18$_{\pm0.77}$} & 45.74$_{\pm0.73}$ & \textbf{46.11$_{\pm0.69}$} \\
 & 90\% & 49.42$_{\pm0.72}$ & \textbf{49.60$_{\pm0.74}$} & 47.41$_{\pm0.84}$ & \textbf{48.06$_{\pm0.71}$} & \textbf{49.04$_{\pm0.79}$} & 48.28$_{\pm0.81}$ & 45.61$_{\pm0.70}$ & \textbf{47.62$_{\pm0.83}$} \\
 & 95\% & \textbf{49.04$_{\pm0.76}$} & 48.43$_{\pm0.78}$ & 45.55$_{\pm0.91}$ & \textbf{45.61$_{\pm0.86}$} & 41.66$_{\pm1.21}$ & \textbf{45.58$_{\pm0.75}$} & 44.53$_{\pm0.84}$ & \textbf{45.14$_{\pm0.72}$} \\
 & 99\% & 40.99$_{\pm1.34}$ & \textbf{41.86$_{\pm1.19}$} & 35.15$_{\pm1.48}$ & \textbf{36.74$_{\pm1.42}$} & 40.79$_{\pm0.94}$ & \textbf{45.77$_{\pm1.07}$} & \textbf{40.85$_{\pm0.99}$} & 37.93$_{\pm1.22}$ \\
\midrule
\multirow{4}{*}{VGG-16}
 & 80\% & 41.93$_{\pm0.82}$ & \textbf{42.84$_{\pm0.85}$} & 40.31$_{\pm0.95}$ & \textbf{41.39$_{\pm0.91}$} & 46.95$_{\pm0.78}$ & \textbf{48.93$_{\pm0.84}$} & 45.78$_{\pm0.89}$ & \textbf{46.17$_{\pm0.75}$} \\
 & 90\% & 41.69$_{\pm0.86}$ & \textbf{42.11$_{\pm0.87}$} & 38.12$_{\pm1.12}$ & \textbf{40.61$_{\pm0.93}$} & \textbf{47.30$_{\pm0.79}$} & 46.28$_{\pm0.76}$ & 44.22$_{\pm0.81}$ & \textbf{46.17$_{\pm0.88}$} \\
 & 95\% & \textbf{40.21$_{\pm0.97}$} & 39.13$_{\pm0.99}$ & 31.81$_{\pm1.42}$ & \textbf{38.03$_{\pm1.08}$} & 46.51$_{\pm0.75}$ & \textbf{47.96$_{\pm0.71}$} & 42.43$_{\pm0.84}$ & \textbf{43.78$_{\pm0.73}$} \\
 & 99\% & 24.22$_{\pm1.52}$ & \textbf{36.41$_{\pm1.21}$} & 20.54$_{\pm1.68}$ & \textbf{29.67$_{\pm1.49}$} & 43.42$_{\pm0.77}$ & \textbf{43.91$_{\pm0.81}$} & 31.06$_{\pm1.34}$ & \textbf{32.64$_{\pm1.41}$} \\
\midrule
\multirow{4}{*}{WRN28-4}
 & 80\% & 50.45$_{\pm0.81}$ & \textbf{50.59$_{\pm0.73}$} & 50.31$_{\pm0.78}$ & \textbf{50.41$_{\pm0.76}$} & 43.79$_{\pm0.74}$ & \textbf{47.02$_{\pm0.78}$} & \textbf{49.43$_{\pm0.73}$} & 47.50$_{\pm0.71}$ \\
 & 90\% & 50.56$_{\pm0.77}$ & \textbf{50.79$_{\pm0.72}$} & 47.75$_{\pm0.88}$ & \textbf{49.30$_{\pm0.80}$} & 45.89$_{\pm0.75}$ & \textbf{46.31$_{\pm0.74}$} & 43.80$_{\pm0.76}$ & \textbf{45.66$_{\pm0.78}$} \\
 & 95\% & 49.07$_{\pm0.91}$ & \textbf{49.37$_{\pm0.87}$} & \textbf{46.97$_{\pm0.97}$} & 46.85$_{\pm0.93}$ & 41.69$_{\pm0.79}$ & \textbf{45.41$_{\pm0.76}$} & 48.01$_{\pm0.75}$ & \textbf{48.35$_{\pm0.73}$} \\
 & 99\% & 38.89$_{\pm1.39}$ & \textbf{39.89$_{\pm1.22}$} & 34.57$_{\pm1.47}$ & \textbf{36.30$_{\pm1.34}$} & \textbf{43.58$_{\pm0.78}$} & 40.87$_{\pm0.81}$ & \textbf{40.57$_{\pm0.79}$} & 38.84$_{\pm0.82}$ \\
\bottomrule
\end{tabular}%
}
\end{table*}

\subsection{Effect of \ssap on Score-Space Sharpness and Mask Stability}\label{sect:exp_sharpness}
We evaluate here the effect of \ssap on the sharpness of the loss landscape parameterized by the importance scores. 
In contrast to conventional approaches, we measure score-space sharpness in the robust loss landscape and adapt the measures accordingly. 
In addition, we introduce the mask stability property to probe the effect of score-space sharpness minimization on mask-search dynamics. We quantify stability via the normalized Hamming distance between the first and subsequent pruning masks and observe that \ssap generally reduces this distance.

\begin{figure}
    \centering
    \includegraphics[width=0.9\linewidth]{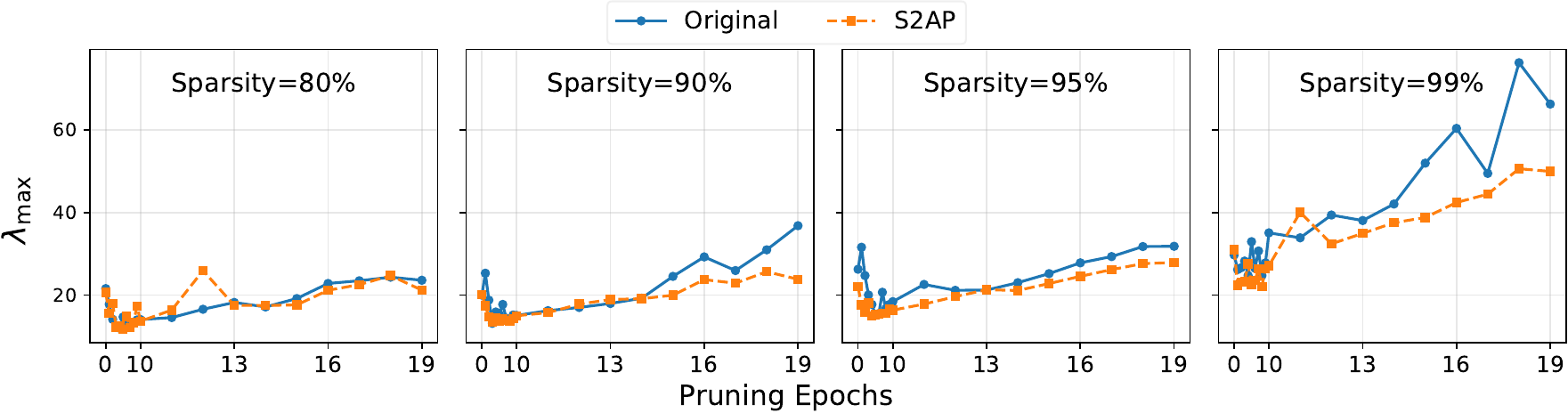}
    \caption{Score-space sharpness measured via largest eigenvalue $\lambda_{max}$ over pruning epochs for HARP on WideResNet28-4 and CIFAR10.}
    \label{fig:eigenvalues_wrn}
\end{figure}

\begin{figure}
    \centering
    \includegraphics[width=0.9\linewidth]{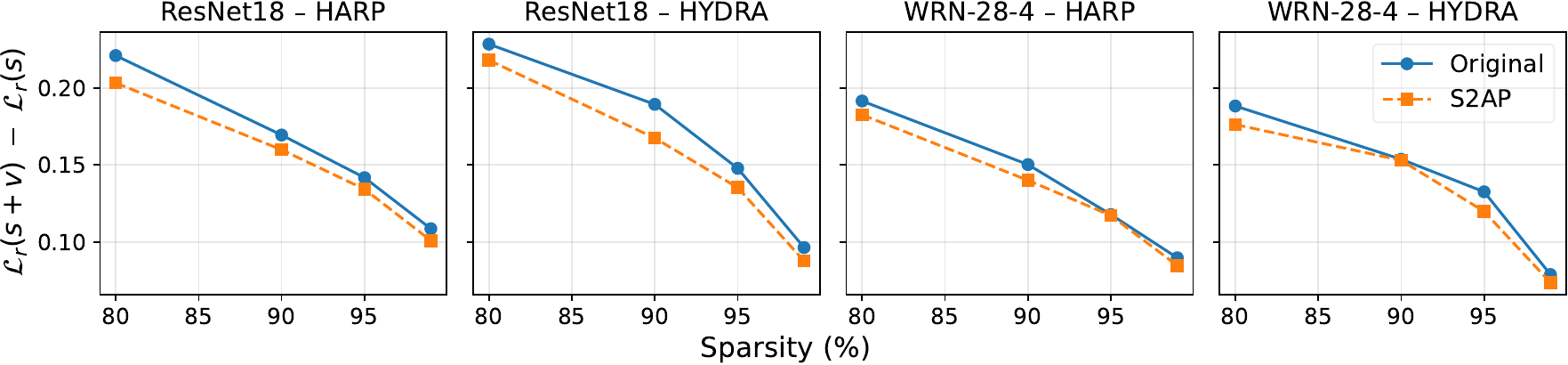}
    \caption{Score-space sharpness measured as difference of perturbed and reference loss values on ResNet18 and WideResNet28-4 CIFAR10 pruned models.}
    \label{fig:sharpness_all}
\end{figure}

\myparagraph{Minimized Score-Space Sharpness.} 
We measure score-space sharpness relying on (i) the score-space largest eigenvalue $\lambda_{max}$ measure~\citep{jastrzkebski2017three}; and (ii) a loss-difference measure addressing the scale-invariance problem of Hessian-based measures~\citep{dinh17_pmlr, kaur23a}. We measure $\lambda_{max}$ on the score space for each iteration and average the values on each epoch to evaluate sharpness. 
We show in~\autoref{fig:intro_eigenvalues} $\lambda_{max}$ for a ResNet18 model on the CIFAR10 dataset and HARP, which reveals how, across different sparsities, Orig. has the largest eigenvalues (\ie, is sharper) than the \ssap version. 
The same trend can be validated in~\autoref{fig:eigenvalues_wrn} for a WideResNet28-4.   
The loss difference instead is computed by crafting a score-space perturbation added to the scores parameterizing a \robustloss minima, and subtracted from the reference \robustloss value, thus extending the approach from~\cite{andriushchenko23_icml, stutz21_iccv} to the score space. 
In this case, we consider the best \robustloss minima found during the pruning mask search, then compute the difference $\hat{\mathcal{L}}(s+\vct v)-\hat{\mathcal{L}}(s)$, where $\vct v$ is a score perturbation crafted through the Auto-PGD (APGD) optimization approach. 
Care must be taken not to conflate this perturbation, added to already optimized scores to simply estimate the loss sharpness, with the one designed in~\autoref{alg:s2ap} added during optimization to induce sharpness.  
As shown in~\autoref{fig:sharpness_all}, the sharpness of our \ssap approach is lower.  
More details on the sharpness measures and additional experiments can be found in~\autoref{app:apdx_eigen} and~\autoref{app:apdx_lossdiff}. 

\begin{figure}[ht]
  \centering
  \begin{subfigure}[t]{0.51\linewidth}
    \centering
    \includegraphics[width=\linewidth]{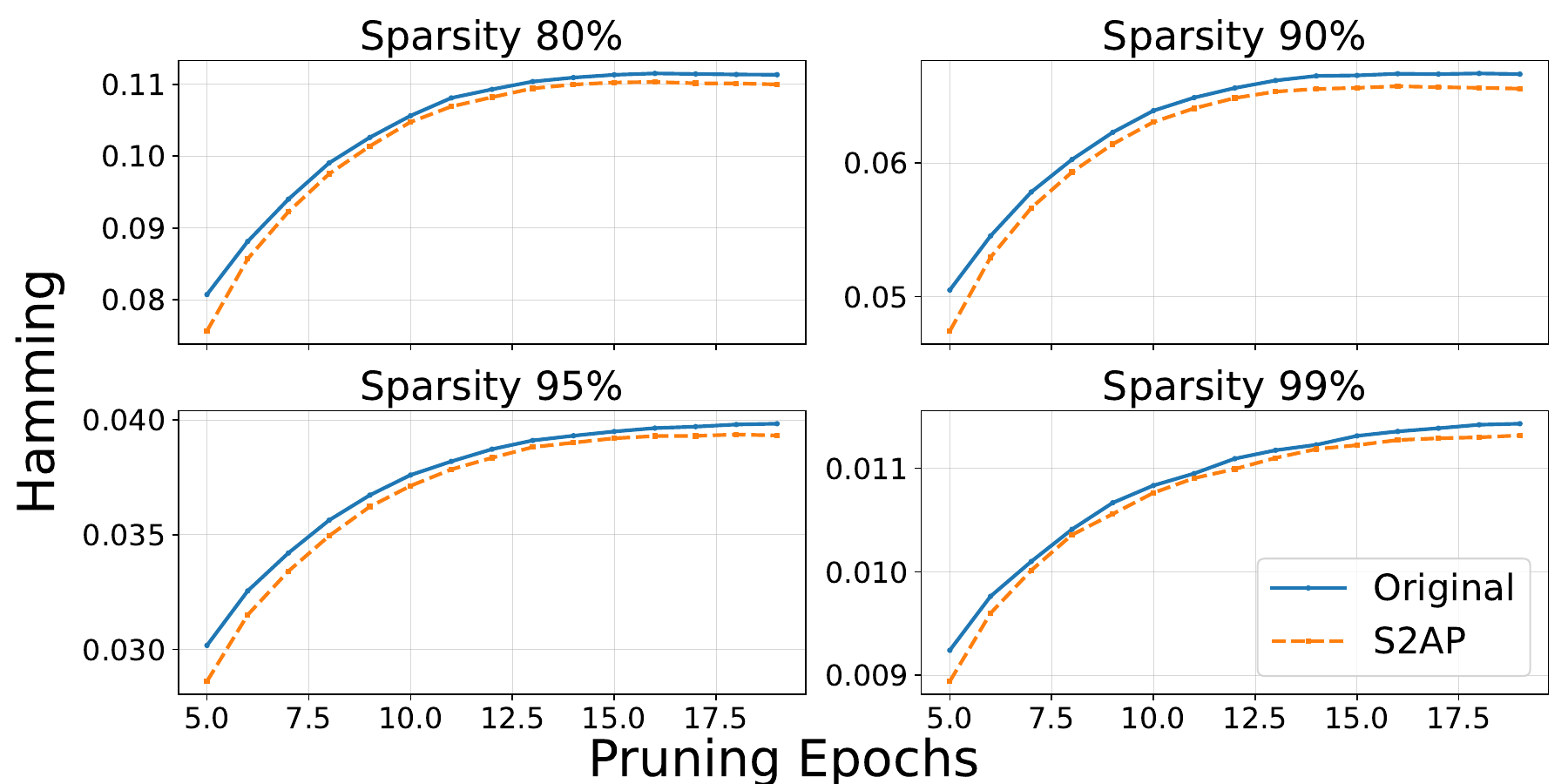}
    \caption{}
    \label{fig:mask_stability_wrn_2x2}
  \end{subfigure}%
  \hspace{0.01\linewidth}
  \begin{subfigure}[t]{0.37\linewidth}
    \centering
    \includegraphics[width=\linewidth]
      {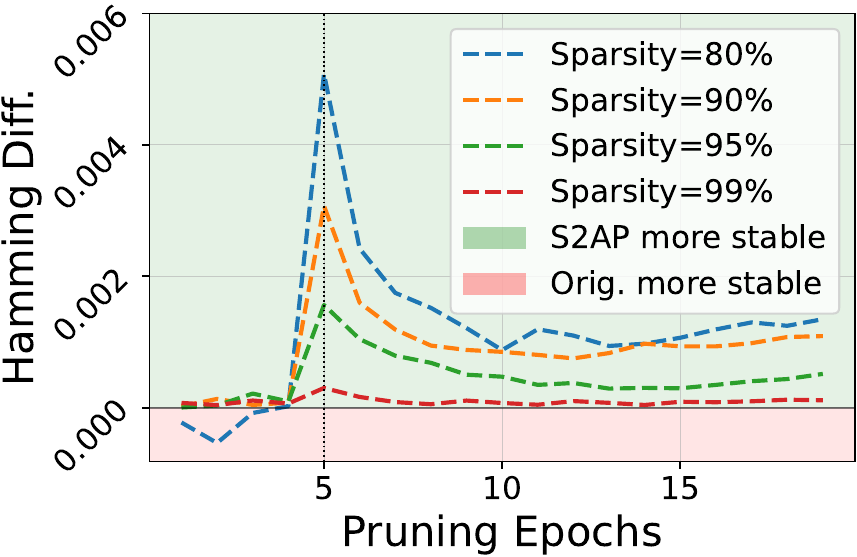}
    \caption{}
    \label{fig:mask_stability_wrn_all}
  \end{subfigure}%
  \caption{The Hamming distance for WideResNet28-4 on CIFAR10. In (a) the single hamming distance from epoch $5$ of \ssap and Orig. HARP. Lower curves indicate higher stability. In (b), the results from the four (a) subplots by subtracting each Original-\ssap curve, thus yielding a positive-green (negative-red) area where \ssap (Original) methods are more stable.}
  \label{fig:wrn_mask_stability}
\end{figure}

\myparagraph{Improved Mask Stability.}
Beyond merely flattening the loss landscape, we study a novel property—\emph{mask stability}—to probe the effect of score-space sharpness minimization on mask-search dynamics.
We aim to test whether a flatter score-space reduces the sensitivity of the selection to small score-variations (\ie., whether the mask search becomes less volatile).
We capture this phenomenon through the normalized Hamming distance~\citep{fu2021drawing}. This allows us to compute the differing $0-1$ values between binary masks $\vct m$. Hence, over the $20$ pruning epochs indexed by $t$, we compute $h = \|\vct m_0 \oplus \vct m_t\|_1/|\vct m_0|$, where $\oplus$ is a XOR operator measuring the differing bits. 
For each pruning epoch, we compute $h_{orig}-h_{\ssap}$, and define a positive region, where \ssap is more stable, and a negative region, where the original method is more stable. 
We show how \ssap improves mask stability for ResNet18 in~\autoref{fig:intro_hamming}, while in~\autoref{fig:mask_stability_wrn_2x2} and~\autoref{fig:mask_stability_wrn_all} we show, respectively, the single Hamming distance curves for original vs. \ssap-based methods and the difference between the curves across all four sparsities. 
Before the five warm-up epochs, being the overall training procedure identical, numerical differences only result in marginal differences between \ssap and the original methods. Then, the spike registered indicates the immediate increased stability induced by \ssap, which denotes how minimizing sharpness makes the mask selection closer to the first computed mask. 
As sparsity increases, since a higher sparsity also implies a lower variability of $ 0$'s and $ 1$'s, the scale of the hamming distance decreases accordingly.   
More details and additional experiments can be found in~\autoref{app:apdx_stability} 

\section{Related Work}\label{sect:related}
\myparagraph{Adversarial Robustness and Sharpness.}
The work from~\cite{wu20_nips} first revealed the correlation between robustness and sharpness. 
In fact, AWP shows that adversarial objectives, such as PGD-AT~\citep{madry18-iclr}, \textit{implicitly} minimize sharpness in the weights' loss landscape. Hence, by \textit{explicitly} minimizing sharpness with respect to both weights and inputs, it improved robustness and flatness. 
On a larger-scale study by~\cite{stutz21_iccv}, and recently also in~\citep{zhang2024duality}, such a relationship has been investigated in more detail and confirmed thoroughly. In our work, we leverage a similar idea to improve the stability and robustness of adversarial pruning methods.

\myparagraph{Pruning and Sharpness.}  
Minimizing sharpness through SAM~\citep{foret21_iclr} has been shown to be beneficial for iterative pruning on BERT models and NLP tasks, compared to the Adam optimizer~\citep{na22_trainflat}. 
The work from~\cite{na22_trainflat} has been extended, besides~\citep{lee2025safe}, to structured pruning and out-of-distribution (OOD) robustness by~\cite{bair24_iclr}. The authors prime the network for pruning based on the rationale that a flatter landscape is more prone to pruning. Hence, they develop an adaptive version of SAM by perturbing the channels more likely to be pruned. 
Further work proposed a single-step sharpness minimization approach aligned with the resource constraints imposed by sparse training~\citep{ji2024single}. 
In contrast, we focus on adversarial robustness (\ie, adversarial pruning) and on score-space sharpness minimization, rather than the typical weights' loss landscape. 
Most importantly, we do not focus on pre-pruning network priming, but rather explicitly operate on score space during the pruning mask search.

From a conceptual perspective, our work is the first to blend the robustness/sharpness/pruning lines of work by proposing a sharpness minimization approach for adversarial pruning. However, we promote the novel concept of score-space sharpness minimization, thus allowing us to measure and improve mask-search stability, besides robustness.

\section{Conclusions, Limitations, and Future Work}\label{sect:conclusion}
We have introduced \ssap, a score-space sharpness minimization for adversarial pruning methods. Leveraging the concept of score-space, \ssap effectively minimizes sharpness, improves the mask-search stability, and consistently increases adversarial robustness across various datasets, models, and sparsities. 
As limitations, we believe that the additional costs of minimizing sharpness, which apply to all standard SAM-like objectives, might be unsustainable in specific application scenarios. Despite being cost minimization out of this work's scope, we believe ``cheaper'' approaches such as the one from~\cite{ji2024single} could be extended to the \ssap case as future work.  
Finally, let us specify how the network architecture choices have been dictated by the availability of state-of-the-art AP methods, which do not extend to more recent transformer architectures. 
Despite being ours, to the best of our knowledge, the first adversarial pruning work considering such architectures, we believe that a consistent setup shift is required for adversarial pruning methods, and hope our work can inspire such improvements. 
To conclude, we remark how \ssap can be extended to any score-based optimization, beyond adversarial pruning.

\myparagraph{Reproducibility Statement.}
We have taken several steps to facilitate reproducibility. The S2AP method is precisely specified in~\autoref{alg:s2ap}; the finetuning objective is given~\autoref{alg:apx_s2ap_finetune}. Our experimental setup---datasets, architectures, sparsity levels, training and evaluation protocols, and threat model---is documented in~\autoref{sect:exp_setup}. Hyperparameter choices are reported in the paper and further discussed in~\autoref{app:appendix_additional_exps}. We describe the score-space sharpness metrics and the mask-stability metric in~\autoref{sect:exp_sharpness} with additional implementation details in~\autoref{app:appendix_sharpness}. In the \emph{supplementary material}, we include an anonymized code archive containing all needed source code, training/evaluation scripts, and the \emph{default configurations} used in our experiments; for transparency, these default settings are also listed throughout the paper where relevant and mirrored in the appendix and configuration files. The code will be publicly released upon acceptance.

\myparagraph{Ethics Statement.}
We do not identify any ethical concerns associated with this work. Our study does not involve human subjects, user interaction, or personally identifiable information. All experiments use standard, publicly available datasets (CIFAR-10, SVHN, ImageNet) under their respective licenses. The proposed method is defensive—focusing on pruning and adversarial robustness—and does not introduce new attack capabilities beyond standard, widely used evaluation protocols (e.g., PGD, AutoAttack). We are not aware of privacy, security, fairness, or legal compliance issues arising from our methodology or experimental setup, and we have no conflicts of interest or sponsorship to declare. We have read and adhere to the ICLR Code of Ethics.

\section*{Acknowledgments}
This work has been partly supported by the EU-funded Horizon Europe projects ELSA (GA no. 101070617), Sec4AI4Sec (GA no. 101120393), and CoEvolution (GA no. 101168560); and by the projects SERICS (PE00000014) and FAIR (PE00000013) under the MUR National Recovery and Resilience Plan funded by the European Union - NextGenerationEU. We gratefully acknowledge funding by the Helmholtz Association (HGF) within topic ``46.23 Engineering Secure Systems''.

\bibliography{biblio}

\begin{thebibliography}{34}
\providecommand{\natexlab}[1]{#1}
\providecommand{\url}[1]{\texttt{#1}}
\expandafter\ifx\csname urlstyle\endcsname\relax
  \providecommand{\doi}[1]{doi: #1}\else
  \providecommand{\doi}{doi: \begingroup \urlstyle{rm}\Url}\fi

\bibitem[Andriushchenko et~al.(2023)Andriushchenko, Croce, M{\"{u}}ller, Hein, and Flammarion]{andriushchenko23_icml}
Maksym Andriushchenko, Francesco Croce, Maximilian M{\"{u}}ller, Matthias Hein, and Nicolas Flammarion.
\newblock A modern look at the relationship between sharpness and generalization.
\newblock In Andreas Krause, Emma Brunskill, Kyunghyun Cho, Barbara Engelhardt, Sivan Sabato, and Jonathan Scarlett (eds.), \emph{International Conference on Machine Learning, {ICML} 2023, 23-29 July 2023, Honolulu, Hawaii, {USA}}, volume 202 of \emph{Proceedings of Machine Learning Research}, pp.\  840--902. {PMLR}, 2023.
\newblock URL \url{https://proceedings.mlr.press/v202/andriushchenko23a.html}.

\bibitem[Bair et~al.(2024)Bair, Yin, Shen, Molchanov, and Alvarez]{bair24_iclr}
Anna Bair, Hongxu Yin, Maying Shen, Pavlo Molchanov, and Jose~M. Alvarez.
\newblock Adaptive sharpness-aware pruning for robust sparse networks.
\newblock In \emph{The Twelfth International Conference on Learning Representations}, 2024.
\newblock URL \url{https://openreview.net/forum?id=QFYVVwiAM8}.

\bibitem[Biggio et~al.(2013)Biggio, Corona, Maiorca, Nelson, \v{S}rndi\'{c}, Laskov, Giacinto, and Roli]{biggio13-ecml}
B.~Biggio, I.~Corona, D.~Maiorca, B.~Nelson, N.~\v{S}rndi\'{c}, P.~Laskov, G.~Giacinto, and F.~Roli.
\newblock Evasion attacks against machine learning at test time.
\newblock In \emph{ECML-PKDD}, 2013.

\bibitem[Blalock et~al.(2020)Blalock, Ortiz, Frankle, and Guttag]{blalock20_mlsys}
Davis~W. Blalock, Jose Javier~Gonzalez Ortiz, Jonathan Frankle, and John~V. Guttag.
\newblock What is the state of neural network pruning?
\newblock In \emph{Proceedings of Machine Learning and Systems 2020, MLSys}, 2020.

\bibitem[Croce \& Hein(2020)Croce and Hein]{croce20_autoattack}
Francesco Croce and Matthias Hein.
\newblock Reliable evaluation of adversarial robustness with an ensemble of diverse parameter-free attacks.
\newblock In \emph{ICML}, 2020.

\bibitem[Deng et~al.(2009)Deng, Dong, Socher, Li, Li, and Fei-Fei]{deng2009imagenet}
Jia Deng, Wei Dong, Richard Socher, Li-Jia Li, Kai Li, and Li~Fei-Fei.
\newblock {ImageNet}: A large-scale hierarchical image database.
\newblock In \emph{2009 IEEE Conference on Computer Vision and Pattern Recognition (CVPR)}, pp.\  248--255. IEEE, 2009.

\bibitem[Dinh et~al.(2017)Dinh, Pascanu, Bengio, and Bengio]{dinh17_pmlr}
Laurent Dinh, Razvan Pascanu, Samy Bengio, and Yoshua Bengio.
\newblock Sharp minima can generalize for deep nets.
\newblock In Doina Precup and Yee~Whye Teh (eds.), \emph{Proceedings of the 34th International Conference on Machine Learning}, volume~70 of \emph{Proceedings of Machine Learning Research}, pp.\  1019--1028. PMLR, 06--11 Aug 2017.
\newblock URL \url{https://proceedings.mlr.press/v70/dinh17b.html}.

\bibitem[Foret et~al.(2021)Foret, Kleiner, Mobahi, and Neyshabur]{foret21_iclr}
Pierre Foret, Ariel Kleiner, Hossein Mobahi, and Behnam Neyshabur.
\newblock Sharpness-aware minimization for efficiently improving generalization.
\newblock In \emph{International Conference on Learning Representations}, 2021.
\newblock URL \url{https://openreview.net/forum?id=6Tm1mposlrM}.

\bibitem[Frankle \& Carbin(2019)Frankle and Carbin]{frankle18_iclr}
Jonathan Frankle and Michael Carbin.
\newblock The lottery ticket hypothesis: Finding sparse, trainable neural networks.
\newblock In \emph{International Conference on Learning Representations}, 2019.
\newblock URL \url{https://openreview.net/forum?id=rJl-b3RcF7}.

\bibitem[Fu et~al.(2021)Fu, Yu, Zhang, Wu, Ouyang, Cox, and Lin]{fu2021drawing}
Yonggan Fu, Qixuan Yu, Yang Zhang, Shang Wu, Xu~Ouyang, David Cox, and Yingyan Lin.
\newblock Drawing robust scratch tickets: Subnetworks with inborn robustness are found within randomly initialized networks.
\newblock \emph{Advances in Neural Information Processing Systems}, 34:\penalty0 13059--13072, 2021.

\bibitem[Han et~al.(2015)Han, Pool, Tran, and Dally]{han15_nips}
Song Han, Jeff Pool, John Tran, and William~J. Dally.
\newblock Learning both weights and connections for efficient neural networks.
\newblock In \emph{NeurIPS}, 2015.

\bibitem[He et~al.(2016)He, Zhang, Ren, and Sun]{he2016deep}
Kaiming He, Xiangyu Zhang, Shaoqing Ren, and Jian Sun.
\newblock Deep residual learning for image recognition.
\newblock In \emph{Proceedings of the IEEE Conference on Computer Vision and Pattern Recognition (CVPR)}, pp.\  770--778, 2016.

\bibitem[Jastrz{\k{e}}bski et~al.(2017)Jastrz{\k{e}}bski, Kenton, Arpit, Ballas, Fischer, Bengio, and Storkey]{jastrzkebski2017three}
Stanis{\l}aw Jastrz{\k{e}}bski, Zachary Kenton, Devansh Arpit, Nicolas Ballas, Asja Fischer, Yoshua Bengio, and Amos Storkey.
\newblock Three factors influencing minima in sgd.
\newblock \emph{arXiv preprint arXiv:1711.04623}, 2017.

\bibitem[Ji et~al.(2024)Ji, Li, Fu, Afghah, Guo, Yuan, and Ma]{ji2024single}
Jie Ji, Gen Li, Jingjing Fu, Fatemeh Afghah, Linke Guo, Xiaoyong Yuan, and Xiaolong Ma.
\newblock A single-step, sharpness-aware minimization is all you need to achieve efficient and accurate sparse training.
\newblock \emph{Advances in Neural Information Processing Systems}, 37:\penalty0 44269--44290, 2024.

\bibitem[Kaur et~al.(2023)Kaur, Cohen, and Lipton]{kaur23a}
Simran Kaur, Jeremy Cohen, and Zachary~Chase Lipton.
\newblock On the maximum hessian eigenvalue and generalization.
\newblock In Javier Antorán, Arno Blaas, Fan Feng, Sahra Ghalebikesabi, Ian Mason, Melanie~F. Pradier, David Rohde, Francisco J.~R. Ruiz, and Aaron Schein (eds.), \emph{Proceedings on "I Can't Believe It's Not Better! - Understanding Deep Learning Through Empirical Falsification" at NeurIPS 2022 Workshops}, volume 187 of \emph{Proceedings of Machine Learning Research}, pp.\  51--65. PMLR, 03 Dec 2023.
\newblock URL \url{https://proceedings.mlr.press/v187/kaur23a.html}.

\bibitem[Krizhevsky et~al.(2009)Krizhevsky, Hinton, et~al.]{krizhevsky2009learning}
Alex Krizhevsky, Geoffrey Hinton, et~al.
\newblock Learning multiple layers of features from tiny images.
\newblock \emph{Technical Report}, 2009.

\bibitem[LeCun et~al.(1989)LeCun, Denker, and Solla]{lecun89_nips}
Yann LeCun, John~S. Denker, and Sara~A. Solla.
\newblock Optimal brain damage.
\newblock In \emph{NIPS}, 1989.

\bibitem[Lee et~al.(2025)Lee, Lee, Chung, and Lee]{lee2025safe}
Dongyeop Lee, Kwanhee Lee, Jinseok Chung, and Namhoon Lee.
\newblock {SAFE}: Finding sparse and flat minima to improve pruning.
\newblock In \emph{Forty-second International Conference on Machine Learning}, 2025.
\newblock URL \url{https://openreview.net/forum?id=10l1pGeOcK}.

\bibitem[Liu \& Wang(2023)Liu and Wang]{liu2023lessons}
Shiwei Liu and Zhangyang Wang.
\newblock Ten lessons we have learned in the new "sparseland": A short handbook for sparse neural network researchers, 2023.

\bibitem[{Madry} et~al.(2018){Madry}, {Makelov}, {Schmidt}, {Tsipras}, and {Vladu}]{madry18-iclr}
A.~{Madry}, A.~{Makelov}, L.~{Schmidt}, D.~{Tsipras}, and A.~{Vladu}.
\newblock Towards deep learning models resistant to adversarial attacks.
\newblock In \emph{ICLR}, 2018.

\bibitem[Na et~al.(2022)Na, Mehta, and Strubell]{na22_trainflat}
Clara Na, Sanket~Vaibhav Mehta, and Emma Strubell.
\newblock {Train Flat, Then Compress: Sharpness-Aware Minimization Learns More Compressible Models}.
\newblock In \emph{Findings of the Conference on Empirical Methods in Natural Language Processing}, 2022.
\newblock URL \url{https://arxiv.org/abs/2205.12694}.

\bibitem[Netzer et~al.(2011)Netzer, Wang, Coates, Bissacco, Wu, and Ng]{netzer2011reading}
Yuval Netzer, Tao Wang, Adam Coates, Antonio Bissacco, Bo~Wu, and Andrew~Y. Ng.
\newblock Reading digits in natural images with unsupervised feature learning.
\newblock In \emph{NIPS Deep Learning and Unsupervised Feature Learning Workshop}, 2011.

\bibitem[Piras et~al.(2024)Piras, Pintor, Demontis, Biggio, Giacinto, and Roli]{piras2024_ap}
Giorgio Piras, Maura Pintor, Ambra Demontis, Battista Biggio, Giorgio Giacinto, and Fabio Roli.
\newblock Adversarial pruning: A survey and benchmark of pruning methods for adversarial robustness.
\newblock \emph{arXiv preprint arXiv:2409.01249}, 2024.

\bibitem[Ramanujan et~al.(2020)Ramanujan, Wortsman, Kembhavi, Farhadi, and Rastegari]{ramanujan20_ste}
Vivek Ramanujan, Mitchell Wortsman, Aniruddha Kembhavi, Ali Farhadi, and Mohammad Rastegari.
\newblock What's hidden in a randomly weighted neural network?
\newblock In \emph{Proceedings of the IEEE/CVF conference on computer vision and pattern recognition}, pp.\  11893--11902, 2020.

\bibitem[Sehwag et~al.(2019)Sehwag, Wang, Mittal, and Jana]{sehwag19_towards}
Vikash Sehwag, Shiqi Wang, Prateek Mittal, and Suman Jana.
\newblock Towards compact and robust deep neural networks.
\newblock \emph{CoRR}, abs/1906.06110, 2019.

\bibitem[Sehwag et~al.(2020)Sehwag, Wang, Mittal, and Jana]{sehwag20_nips}
Vikash Sehwag, Shiqi Wang, Prateek Mittal, and Suman Jana.
\newblock {HYDRA:} pruning adversarially robust neural networks.
\newblock In \emph{NeurIPS}, 2020.

\bibitem[Simonyan \& Zisserman(2015)Simonyan and Zisserman]{simonyan2015very}
Karen Simonyan and Andrew Zisserman.
\newblock Very deep convolutional networks for large-scale image recognition.
\newblock In \emph{International Conference on Learning Representations (ICLR)}, 2015.
\newblock arXiv:1409.1556.

\bibitem[Stutz et~al.(2021)Stutz, Hein, and Schiele]{stutz21_iccv}
David Stutz, Matthias Hein, and Bernt Schiele.
\newblock Relating adversarially robust generalization to flat minima.
\newblock In \emph{Proceedings of the IEEE/CVF International Conference on Computer Vision}, pp.\  7807--7817, 2021.

\bibitem[Szegedy et~al.(2014)Szegedy, Zaremba, Sutskever, Bruna, Erhan, Goodfellow, and Fergus]{szegedy14-iclr}
Christian Szegedy, Wojciech Zaremba, Ilya Sutskever, Joan Bruna, Dumitru Erhan, Ian Goodfellow, and Rob Fergus.
\newblock Intriguing properties of neural networks.
\newblock In \emph{International Conference on Learning Representations}, 2014.

\bibitem[Wu et~al.(2020)Wu, Xia, and Wang]{wu20_nips}
Dongxian Wu, Shu-Tao Xia, and Yisen Wang.
\newblock Adversarial weight perturbation helps robust generalization.
\newblock \emph{Advances in neural information processing systems}, 33:\penalty0 2958--2969, 2020.

\bibitem[Zagoruyko \& Komodakis(2016)Zagoruyko and Komodakis]{zagoruyko2016wide}
Sergey Zagoruyko and Nikos Komodakis.
\newblock Wide residual networks.
\newblock In \emph{British Machine Vision Conference (BMVC)}, pp.\  87.1--87.12. BMVA Press, 2016.

\bibitem[Zhang et~al.(2019)Zhang, Yu, Jiao, Xing, Ghaoui, and Jordan]{zhang19_icml}
Hongyang Zhang, Yaodong Yu, Jiantao Jiao, Eric~P. Xing, Laurent~El Ghaoui, and Michael~I. Jordan.
\newblock Theoretically principled trade-off between robustness and accuracy.
\newblock In \emph{ICML}, 2019.

\bibitem[Zhang et~al.(2024)Zhang, He, Zhu, Chen, Wang, and Wei]{zhang2024duality}
Yihao Zhang, Hangzhou He, Jingyu Zhu, Huanran Chen, Yifei Wang, and Zeming Wei.
\newblock On the duality between sharpness-aware minimization and adversarial training.
\newblock \emph{arXiv preprint arXiv:2402.15152}, 2024.

\bibitem[Zhao \& Wressnegger(2023)Zhao and Wressnegger]{zhao23_iclr}
Qi~Zhao and Christian Wressnegger.
\newblock Holistic adversarially robust pruning.
\newblock In \emph{The Eleventh International Conference on Learning Representations}, 2023.
\newblock URL \url{https://openreview.net/forum?id=sAJDi9lD06L}.

\end{thebibliography}
\bibliographystyle{utils/iclr2026_conference}

\clearpage
\appendix
\section*{\LARGE \bfseries Supplementary Material for \ssap: Score-space Sharpness Minimization for Adversarial Pruning}
The supplementary material is organized as follows: 
\begin{itemize}
    \item \textbf{\autoref{app:appendix_ssap}}: We discuss additional details for the \ssap method, including pretraining and finetuning details, hyperparameter selection, and overhead computing. 
    \item \textbf{\autoref{app:appendix_additional_exps}}: We show additional experiments validating the applicability and effectiveness of \ssap outside the main testbed, including structured pruning, clean standard accuracy, and robustness to corrupted images. We conclude by discussing and showing the comparison of weights and score perturbations during the pruning stage. 
    \item \textbf{\autoref{app:appendix_sharpness}}: We provide additional details and experiments for the eigenvalue computation, the loss difference measuring sharpness, and the mask stability and hamming distance measure. 
\end{itemize}

\section{Additional \ssap Details.}\label{app:appendix_ssap}
This section describes the additional details concerning our \ssap implementation and results. 
In detail, we first show the results from the pretrained models used in~\autoref{tab:s2ap_cifar10_robust_clean}, \autoref{tab:s2ap_svhn_robust_clean}, \autoref{tab:s2ap_imagenet_robust_clean}, and \autoref{tab:all_bra}. 
Then, we discuss in detail the \ssap finetuning algorithm, which concerns perturbing the remaining sparse weight parameterization $\vct w \odot \vct m^*$ as in~\autoref{eq:SSAP_finetuning}.
We conclude by motivating the choices of the $\gamma$ values bounding the score-perturbations listed in~\autoref{sect:exp_setup}, and computing the overhead induced by our \ssap approach compared to a standard score-based pruning optimization. 

Let us finally specify that the \ssap \textbf{code implementation} is part of the supplementary material and will be publicly released upon paper acceptance.

\subsection{\ssap Pretraining and ImageNet details}\label{app:apdx_pretraining}
\begin{wraptable}{r}{0.35\textwidth}
\centering
\vspace{-10pt} 
\caption{Pretrained models' clean/robust accuracy.}
\label{tab:apx_pretrained}
\resizebox{0.35\textwidth}{!}{%
\begin{tabular}{@{}llc@{}}
\toprule
Model & Dataset & Orig.\\
\midrule
\multirow{2}{*}{ResNet18} 
& CIFAR10 & 81.55 / 49.36  \\
& SVHN & 90.70 / 42.08 \\
\midrule
\multirow{2}{*}{VGG16} 
& CIFAR10 & 80.18 / 45.09  \\
& SVHN & 89.41 / 45.71 \\
\midrule
\multirow{2}{*}{WRN28-4} 
& CIFAR10 & 83.68 / 50.12  \\
& SVHN & 93.23 / 42.35 \\
\midrule
\multirow{1}{*}{ResNet50} 
& ImageNet & 60.25 / 36.82 \\
\bottomrule
\end{tabular}%
}
\vspace{-30pt} 
\end{wraptable}

We pretrain each CIFAR10 and SVHN model using $100$ epochs, and show the resulting adversarial robustness in~\autoref{tab:apx_pretrained}. 
For ImageNet, however, we use the pretrained model provided by~\cite{zhao23_iclr}, and prune for $10$ epochs (of which $5$ warm-up and $5$ \ssap) and finetune for $25$ using the Fast Adversarial Training approach.

\subsection{\ssap Finetuning}\label{app:apdx_finetuning}  
We defined the overall finetuning objective in~\autoref{eq:SSAP_finetuning} as: 
\begin{align}
\label{eq:apx_SSAP_finetuning}\vct w^* \in \underset{\vct w}{\arg \min} \: \underset{\vct \nu}{\max} \:\mathcal{L}_r((\vct w+\vct \nu) \odot \vct m^*) \, ,\\
\label{eq:apx_gamma_bound2} \text{where } \|\vct \nu_l\| \leq \gamma \|\vct w_l\|,
\end{align} 
and $\gamma$ bounds the layer-wise perturbation and scales it based on each layer's weight magnitude, similarly to~\cite{wu20_nips}.
Hence, given the sparse parameterization defined by the mask $\vct m^*$ found during \ssap pruning in~\autoref{alg:s2ap}, the \ssap finetuning formulation of~\autoref{eq:apx_SSAP_finetuning} amounts to perturbing and updating only the non-zero (\ie, non-pruned) weights. 
While the \ssap procedure allows improving sharpness, stability, and robustness of the pruning mask per se, such a procedure enables aligning the finetuning objective with the pruning one and further improves robustness.

We provide a detailed implementation of the finetuning algorithm in~\autoref{alg:apx_s2ap_finetune}.
Overall, the algorithm structure remains similar to~\autoref{alg:s2ap}, with the only major variation that the perturbation $\vct \nu$ is applied on the non-zero weights $\vct w\odot\vct m^*$ only, instead of the entire score-space parameterized by $\vct s$. 
\begin{algorithm}[t]
 \SetKwInOut{Input}{Input}
 \SetKwInOut{Output}{Output}
 \DontPrintSemicolon
 \caption{Score-Sharpness-aware Adversarial Finetuning (S2AP Finetune).}
 \label{alg:apx_s2ap_finetune}
 \Input{$\vct w \in \mathbb{R}^p$, pretrained weights; $\vct m^* \in \{0,1\}^p$, binary pruning mask; $\vct x$, training input samples; $\eta$, learning rate; $I$, number of iterations; $L$, number of layers; $\gamma$, perturbation scaling factor; $\hat{\mathcal{L}}$, robust loss.} 
 \Output{Finetuned weights $\vct w^* \in \mathbb{R}^p$}

 Initialize $\vct \nu \gets 0$\;

 \For{$i \gets 1$ \KwTo $I$} {
   Generate adversarial examples on pruned model $\vct x_i' \gets \vct x_i + \vct \delta_i$\;
   Compute robust loss $\hat{\mathcal{L}}(\vct w\odot \vct m^*) = \hat{\mathcal{L}}(\vct w\odot \vct m^*, \mathcal{D})$\;

   Perturb pruned weights $\vct \nu \gets \vct \nu + \eta \left( \nabla_{\nu}\hat{\mathcal{L}}((\vct w + \vct \nu) \odot \vct m^*) / \|\nabla_{\nu}\hat{\mathcal{L}}((\vct w + \vct \nu) \odot \vct m^*)\| \right)$\;

   \For{$l \gets 1$ \KwTo $L$} {
     \If{$\|\vct \nu^{(l)}\| > \gamma \,\|\vct w^{(l)}\|$} {
       Project $\vct \nu^{(l)} \gets \left( \gamma \|\vct w^{(l)}\| / \|\vct \nu^{(l)}\| \right)\vct \nu^{(l)}$\;
     }
   }

   Update weights: $\vct w \gets \vct w - \eta \left( \nabla_{\vct w} \hat{\mathcal{L}}((\vct w + \vct \nu) \odot \vct m^*) / \|\nabla_{\vct w} \hat{\mathcal{L}}((\vct w + \vct \nu) \odot \vct m^*)\| \right)$\;

   Restore weights: $\vct w \gets \vct w - \vct \nu$\;
 }

 \textbf{return} $\vct w^* \gets \vct w$
\end{algorithm}

\subsection{\texorpdfstring{$\gamma$}{gamma}-Selection}\label{app:apdx_gamma} 
We select the $\gamma$ values, bounding the perturbation during \ssap pruning and finetuning, based on the adversarial robustness achieved choosing among a set of values $\gamma=\{0.00075, 0.001, 0.0025, 0.005, 0.0075, 0.01\}$.
\begin{figure}[ht]
    \centering
    \includegraphics[width=0.99\linewidth]{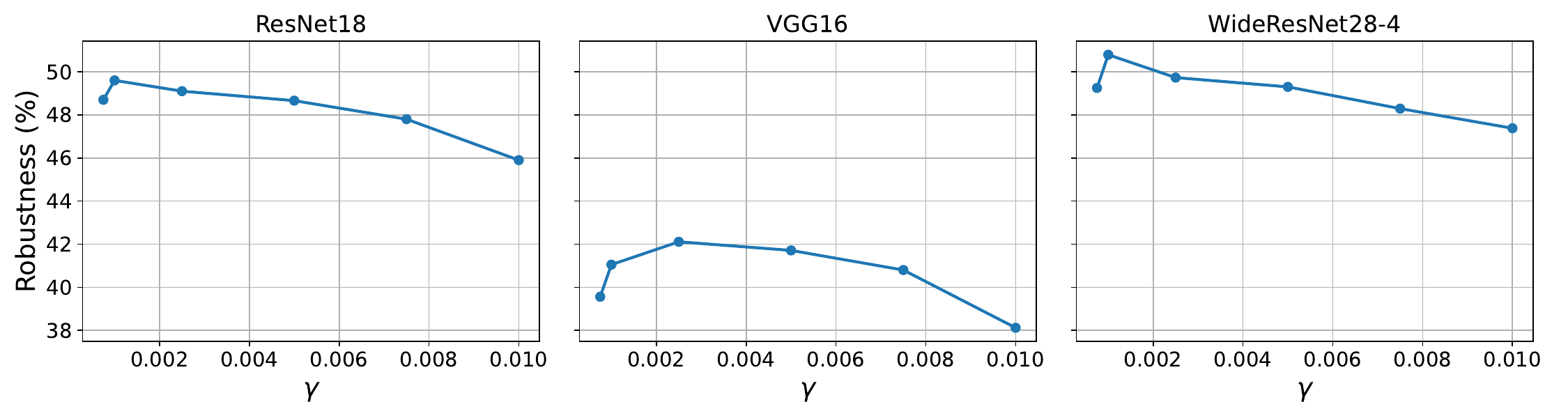}
    \caption{Robustness of \ssap pruning masks found using different $\gamma$ values bounding the score perturbation.}
    \label{fig:apx_gammarobustness}
\end{figure} 
We show in~\autoref{fig:apx_gammarobustness} the gamma search results for the CIFAR10 dataset, HARP method~\cite{zhao23_iclr} at 90\% sparsity. 
We repeat such an evaluation for each model/dataset combination at such sparsity, which we find descriptive of the trend on different sparsities as well, and find the best $\gamma$ value.
Typically, we see a robustness increase for values prior to the best $\gamma$ found for the models (in this case $0.001$ for ResNet18 and WideResNet28-4, and $0.0025$ for VGG16), and then a corresponding robustness decrease after the best found $\gamma$.

\begin{table}[ht]
\centering
\caption{\ssap overhead computation. We compute the time (hrs) required on a NVIDIA RTX A600 for each model/dataset combination, and report the average time required on different sparsities.}
\label{tab:apx_overhead}
\resizebox{0.65\textwidth}{!}{%
\begin{tabular}{@{}llccc@{}}
\toprule
Model & Dataset & Orig. (hrs) & \ssap (hrs) & Overhead (\%)\\
\midrule
\multirow{2}{*}{ResNet18} 
& CIFAR10 & 3.27 & 3.96 & 17.42\% \\
& SVHN & 4.91 & 5.21 & 5.75\% \\
\midrule
\multirow{2}{*}{VGG16} 
& CIFAR10 & 1.41 & 1.73 & 18.49\% \\
& SVHN & 2.43 & 2.75 & 11.63\% \\
\midrule
\multirow{2}{*}{WRN28-4} 
& CIFAR10 & 6.12 & 6.97 & 12.20\% \\
& SVHN & 6.77 & 7.31 & 7.38\% \\
\midrule
\multirow{1}{*}{ResNet-50} 
& ImageNet & 15.08 & 17.11 & 13.46\% \\
\bottomrule
\end{tabular}%
}
\end{table}
 
\subsection{\ssap Computational Overhead}\label{app:apdx_overhead} 
The \ssap procedure of~\autoref{alg:s2ap} inevitably induces a computational overhead.
To provide an estimate of the required overhead, we report in~\autoref{tab:apx_overhead} the time required by the original pruning methods (Orig.) and \ssap versions during pruning and average over the four sparsities.
All experiments were conducted on a machine equipped with three NVIDIA RTX A6000 GPUs (48GB each), and the results of~\autoref{tab:apx_overhead} were conducted on one of these 3 GPUs. 
Specifically, we report in~\autoref{tab:apx_overhead} the results for CIFAR10 and SVHN models on $20$ epochs ($5$ epochs for ImageNet) and batch size $128$ without warm-up, thus allowing an equal comparison of original and \ssap procedures. Generally, we see an average increase in computing time of $15\%$ circa, which, while it might be negligible in some application scenarios, still increases the overall computation.

\section{Additional Experiments}\label{app:appendix_additional_exps}
We discuss here the additional experiments for \ssap. Precisely, we extend our approach to structured pruning, a standard ``clean'' pruning task, compare \ssap with AWP during the pruning stage, and finally analyze the effectiveness of \ssap on the common corruptions dataset.

\begin{table*}[ht]
\centering
\caption{CIFAR-10 and SVHN results using RLTH with ResNet18, VGG-16, and WideResNet-28-4 across sparsity. Each cell shows clean/robust$_{\pm{std}}$ accuracy and the difference between Orig. and \ssap\ robust generalization gap ($\Delta$). In bold, the model with the highest robustness.}
\label{tab:s2ap_rlth_cifar10_svhn}
\resizebox{\textwidth}{!}{%
\begin{tabular}{@{}llcccccc@{}}
\toprule
\multirow{2}{*}{Network} & \multirow{2}{*}{Sparsity} &
\multicolumn{3}{c}{CIFAR-10 (RLTH)} &
\multicolumn{3}{c}{SVHN (RLTH)} \\
\cmidrule(lr){3-5} \cmidrule(l){6-8}
& & Orig. & \ssap & $\Delta$ & Orig. & \ssap & $\Delta$ \\
\midrule
\multirow{4}{*}{ResNet18}
& 80\% & 67.72 / 33.58 & \textbf{68.13 / 33.80} & +0.37 & 85.02 / 44.60 & \textbf{84.13} / \textbf{44.66} & +0.95 \\
& 90\% & 69.32 / 34.42 & \textbf{69.30 / 34.92} & +0.52 & 83.65 / 44.07 & \textbf{84.51 / 44.50} & +1.29 \\
& 95\% & 68.56 / 34.90 & \textbf{69.93 / 35.38} & +1.05 & 84.83 / 42.78 & \textbf{84.51} / \textbf{43.50} & +1.04 \\
& 99\% & \textbf{66.19 / 32.66} & 60.27 / 31.09 & -4.45 & 81.72 / 41.59 & \textbf{80.61} / \textbf{41.73} & +1.25 \\
\midrule
\multirow{4}{*}{VGG16}
& 80\% & 18.63 / 11.04 & \textbf{22.62 / 12.20} & +1.19 & 32.89 / 18.70 & \textbf{32.83 / 19.01} & +0.37 \\
& 90\% & 23.36 / 13.17 & \textbf{24.63 / 13.19} & +1.01 & \textbf{34.21 / 18.78} & 37.20 / 17.31 & -1.92 \\
& 95\% & 30.04 / 12.06 & \textbf{26.62 / 19.33} & +6.69 & \textbf{37.29 / 20.06} & 34.40 / \textbf{21.86} & +2.71 \\
& 99\% & \textbf{18.36 / 14.47} & 19.09 / 12.50 & -1.70 & \textbf{20.06 / 19.68} & 21.68 / 18.00 & -2.30 \\
\midrule
\multirow{4}{*}{WRN28-4}
& 80\% & 68.94 / 34.55 & \textbf{69.71 / 34.91} & +0.65 & 87.82 / 44.52 & \textbf{87.81 / 44.71} & +0.36 \\
& 90\% & 70.05 / 33.53 & \textbf{69.65 / 34.39} & +0.86 & 88.83 / 43.95 & \textbf{85.97 / 44.53} & +1.58 \\
& 95\% & \textbf{69.29 / 34.40} & 68.69 / 33.55 & -0.75 & \textbf{86.82 / 43.85} & 88.13 / 42.95 & -1.56 \\
& 99\% & 63.19 / 29.56 & \textbf{62.13 / 29.83} & +0.89 & 77.29 / 36.68 & \textbf{80.80 / 37.96} & +3.79 \\
\bottomrule
\end{tabular}%
}
\end{table*}

\subsection{Experiments on RLTH}\label{app:apdx_rlth}
As in~\autoref{tab:s2ap_cifar10_robust_clean} and~\autoref{tab:s2ap_svhn_robust_clean}, for the HARP and HYDRA methods, RLTH can benefit from robustness increases from the \ssap method, as we show in~\autoref{tab:s2ap_rlth_cifar10_svhn}. This result is not obvious, as RLTH involves a different pruning pipeline than existing methods. As opposed to starting from a pretrained model, pruning, and then finetuning, such method in fact follows the lottery ticket hypothesis~\cite{frankle18_iclr}, which admits the existence of subnetworks within dense, randomly initialized models. Overall, compared to other methods, we see RLTH pruned models having lower accuracies due to the pruned random initialization and absence of finetuning. The improved robustness of \ssap, considering the absence of finetuning on RLTH, further corroborates to the ablation study discussed in~\autoref{tab:all_bra}, which shows how \ssap, independently from finetuning at all, is capable of reaching higher adversarial robustness from pruning already. 

\begin{table}[t]
\centering
\caption{Channel Pruning with S2AP on CIFAR10 dataset.}
\setlength{\tabcolsep}{6pt}
\begin{tabular}{llcccc}
\toprule
Network & Sparsity (\%) & HARP-Orig. & S2AP-HARP & HYDRA-Orig. & S2AP-HYDRA \\
\midrule
\multirow{2}{*}{ResNet18}
& 4  & 49.60 & \textbf{50.36} & 49.32 & \textbf{50.85} \\
& 15 & 48.28 & \textbf{48.63} & 38.69 & \textbf{39.79} \\
\midrule
\multirow{2}{*}{VGG-16}
& 4  & 47.37 & \textbf{48.18} & 47.02 & 47.33 \\
& 15 & \textbf{37.38} & 37.17 & 33.15 & \textbf{34.53} \\
\bottomrule
\end{tabular}
\label{tab:structured}
\end{table}

\subsection{Experiments on Structured Pruning}\label{app:apdx_structured} 
Unstructured pruning serves as a great mathematical prototype for neural networks, allowing for single weights to be pruned. Empirically, this is widely accepted as an upper-bound on the other important category of pruning methods, \ie, \textit{structured pruning}~\cite{liu2023lessons}. From a practical perspective, structured pruning allows for removing entire network structures, such as channels and filters, and constitutes a readily usable network size reduction. In fact, while unstructured pruning requires a still maturing dedicated hardware, structured pruning implies reducing network size and leveraging it directly~\cite{liu2023lessons}. 
To validate the effectiveness of our \ssap method, given the high relevance of structured pruning methods, we extend, in~\autoref{tab:structured}, experiments of both HARP and HYDRA methods to channel pruning, relying on the ResNet18 and VGG16 networks on CIFAR10 as a testbed. 
Instead of the classic sparsity rate $k$, for channel pruning we refer to the reduction in floating point operations (FLOPs). Specifically, we obtain $4$ or $15$ times fewer FLOPs than the original dense model, thus improving the overall model efficiency and computing time. 
Such a form of sparsity is more compatible with standard hardware acceleration and better suited for real-world deployment. Overall, these results confirm that \ssap generalizes effectively also to different kinds of pruning structures, further reinforcing the versatility of our approach.

\begin{table}[t]
\caption{Clean accuracy (\%) under different sparsity levels. For each pruning method (HARP/HYDRA), we report Orig. and S2AP variants. Bold indicates the best between Orig. and S2AP.}
\label{tab:sparsity-clean-accuracy}
\centering
\setlength{\tabcolsep}{6pt}
\begin{tabular}{llcccc}
\toprule
Network & Sparsity (\%) & HARP-Orig. & S2AP-HARP & HYDRA-Orig. & S2AP-HYDRA \\
\midrule
\multirow{4}{*}{ResNet18}
& 80 & 94.70 & \textbf{94.85} & \textbf{94.90} & 94.61 \\
& 90 & 94.12 & \textbf{94.89} & 94.37 & \textbf{94.73} \\
& 95 & 93.18 & \textbf{94.56} & 94.20 & \textbf{94.84} \\
& 99 & 92.27 & \textbf{93.01} & 90.22 & \textbf{90.38} \\
\midrule
\multirow{4}{*}{VGG-16}
& 80 & 92.17 & \textbf{92.82} & 92.46 & \textbf{93.20} \\
& 90 & 92.34 & \textbf{92.99} & 92.52 & \textbf{93.70} \\
& 95 & 92.41 & \textbf{93.03} & 91.41 & \textbf{91.95} \\
& 99 & \textbf{90.96} & 91.76 & 87.32 & \textbf{87.40} \\
\bottomrule
\end{tabular}
\end{table}

\subsection{Experiments on Standard Clean Pruning}\label{app:apdx_clean} 
On several occasions throughout the paper, we remarked on the generality of the \ssap method beyond the specific adversarial pruning task. 
We thus aim to first confirm the \ssap effectiveness and utility on the most basic task required by such networks: standard classification. Hence, we prune networks using a standard cross-entropy loss, disregarding the adversarial robustness objective, and fine-tune accordingly. 
We show the results of such experiments in~\autoref{tab:sparsity-clean-accuracy}, where we reveal how \ssap improves not only adversarial robustness, but also clean accuracy on a standard classification task for the CIFAR10 dataset. 
We thus confirm the initial claim of general use and applicability of \ssap to different tasks and scenarios, not limited to the adversarial pruning case.

\begin{table}[t]
\caption{Robust accuracy (\%) on CIFAR-10-C under different sparsity levels. Bold indicates the best between Orig. and S2AP for each pruning method.}
\label{tab:cifar10c-robustness}
\centering
\setlength{\tabcolsep}{6pt}
\begin{tabular}{llcccc}
\toprule
Network & Sparsity (\%) & HARP-Orig. & S2AP-HARP & HYDRA-Orig. & S2AP-HYDRA \\
\midrule
\multirow{4}{*}{ResNet18}
& 80 & 72.52 & \textbf{73.08} & 71.75 & \textbf{72.01} \\
& 90 & 72.62 & \textbf{73.12} & 71.54 & \textbf{72.16} \\
& 95 & 72.27 & \textbf{73.23} & 70.02 & \textbf{70.59} \\
& 99 & \textbf{68.52} & 68.48 & 65.41 & \textbf{66.50} \\
\midrule
\multirow{4}{*}{VGG-16}
& 80 & 70.07 & \textbf{70.97} & 68.84 & \textbf{68.98} \\
& 90 & 71.15 & \textbf{71.34} & 69.23 & \textbf{68.71} \\
& 95 & 69.97 & \textbf{70.05} & 68.15 & \textbf{68.33} \\
& 99 & \textbf{66.89} & 67.45 & 59.09 & \textbf{59.26} \\
\midrule
\multirow{4}{*}{WRN}
& 80 & 72.73 & \textbf{72.88} & 72.59 & \textbf{73.54} \\
& 90 & 72.54 & \textbf{73.06} & 71.75 & \textbf{73.08} \\
& 95 & 73.03 & \textbf{73.32} & 72.83 & \textbf{72.85} \\
& 99 & 67.63 & \textbf{67.95} & 65.61 & \textbf{66.04} \\
\bottomrule
\end{tabular}
\end{table}

\subsection{Experiments on Corruptions}\label{app:apdx_corruptions} 
Following on from the previous experiments, extending to standard pruning, it is likewise relevant to consider further tasks. 
We thus choose to test on the general robustness to corruption task by including experiments on the CIFAR10-C dataset. 
We select a corruption severity of $3$, and show the results in~\autoref{tab:cifar10c-robustness}. 
As in previous experiments, we demonstrate how \ssap is further applicable to different tasks and keeps its superiority compared to other methods. We thus believe that such an extension corroborates the claims and results obtained in adversarial robustness, besides broadening the method's applicability.

\begin{table}[t]
\caption{ResNet18 on CIFAR-10: accuracy (\%) under different sparsity levels when pruning with AWP (perturbing weights) vs. S2AP (perturbing scores). Bold indicates the best between AWP and S2AP for each method.}
\label{tab:resnet18-awp-s2ap}
\centering
\setlength{\tabcolsep}{6pt}
\begin{tabular}{llcccc}
\toprule
Network & Sparsity (\%) & HARP-AWP & HARP-S2AP & HYDRA-AWP & HYDRA-S2AP \\
\midrule
\multirow{4}{*}{ResNet18}
& 80 & 47.32 & \textbf{49.55} & 46.12 & \textbf{48.98} \\
& 90 & 47.80 & \textbf{49.60} & 45.19 & \textbf{48.06} \\
& 95 & 46.91 & \textbf{48.43} & 42.77 & \textbf{45.61} \\
& 99 & 40.35 & \textbf{41.86} & 34.34 & \textbf{36.74} \\
\bottomrule
\end{tabular}
\end{table}

\subsection{Perturbing Weights or Scores?}\label{app:apdx_awp_vs_ssap} 
One of the big novelties that can be found in \ssap is the focus on the score-space, rather than the usual weight-space where prior sharpness-minimization approaches focused in the past. 
In turn, a natural question is whether sharpness minimization should be performed in weight space, as done in prior work such as Adversarial Weight Perturbations (AWP), or in score space, as we propose in \ssap. 
In adversarial pruning, the pruning mask is determined by the ranking of importance scores rather than the weights themselves. 
Hence, perturbing scores directly addresses the variables that drive mask selection, potentially stabilizing the top-k cutoff. While this intuition suggests a better alignment with the pruning objective, our main justification is empirical. 
As shown in~\autoref{tab:resnet18-awp-s2ap}, perturbing scores during mask search consistently leads to higher robust accuracy than perturbing weights, across different networks and datasets. These results, which indicate the mask robustness before finetuning as in~\autoref{tab:all_bra}, indicate that score-space perturbations are more effective at preserving robustness in adversarial pruning than their weight-space counterparts.
While a more formal reason describing the differences between applying AWP or \ssap during pruning is missing, we believe that a role behind the greater success of score perturbations could also be played by the increased mask stability.

\section{Measuring Score-Space Sharpness and Mask Stability}\label{app:appendix_sharpness}
We measure score-space sharpness relying on two specific approaches: the largest eigenvalue computation $\lambda_{max}$ and the loss difference (following~\cite{stutz21_iccv, andriushchenko23_icml}). 
We dedicate this section to describing both approaches in detail, and provide additional experiments and results on more model and dataset combinations. 
In addition to minimizing sharpness, however, \ssap also improves the mask stability during pruning. 
In turn, we conclude this section by describing the proposed measure in detail and showing additional experiments. 

\subsection{Measuring Largest Eigenvalue}\label{app:apdx_eigen}  
\begin{figure}
    \centering
    \includegraphics[width=1\linewidth]{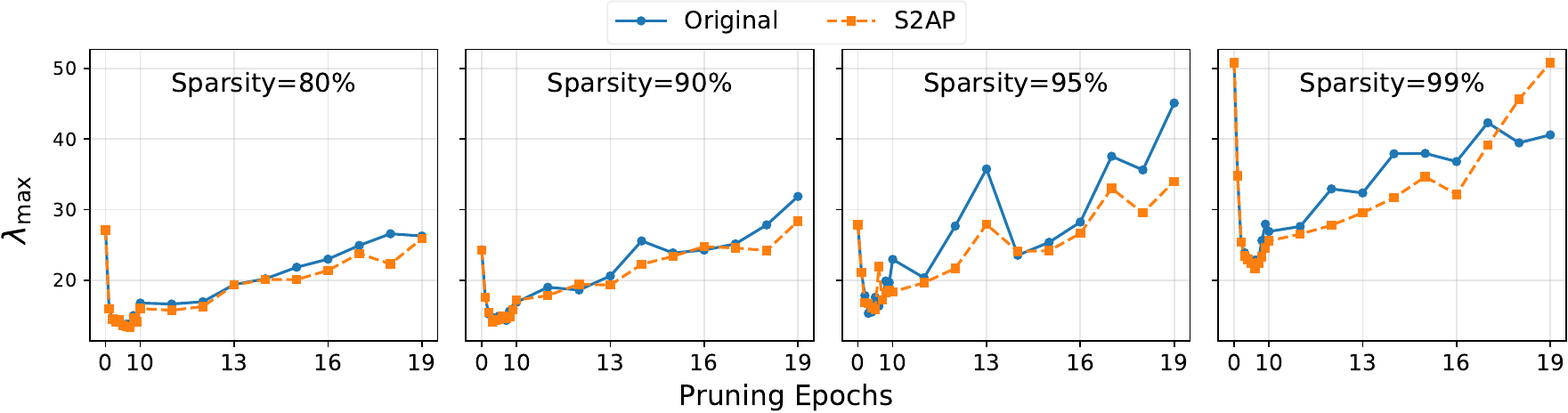}
    \caption{Largest eigenvalue across HYDRA pruning epochs for ResNet18 on CIFAR10.}
    \label{fig:apx_eigen_r18_c10_score}
\end{figure}

\begin{figure}
    \centering
    \includegraphics[width=1\linewidth]{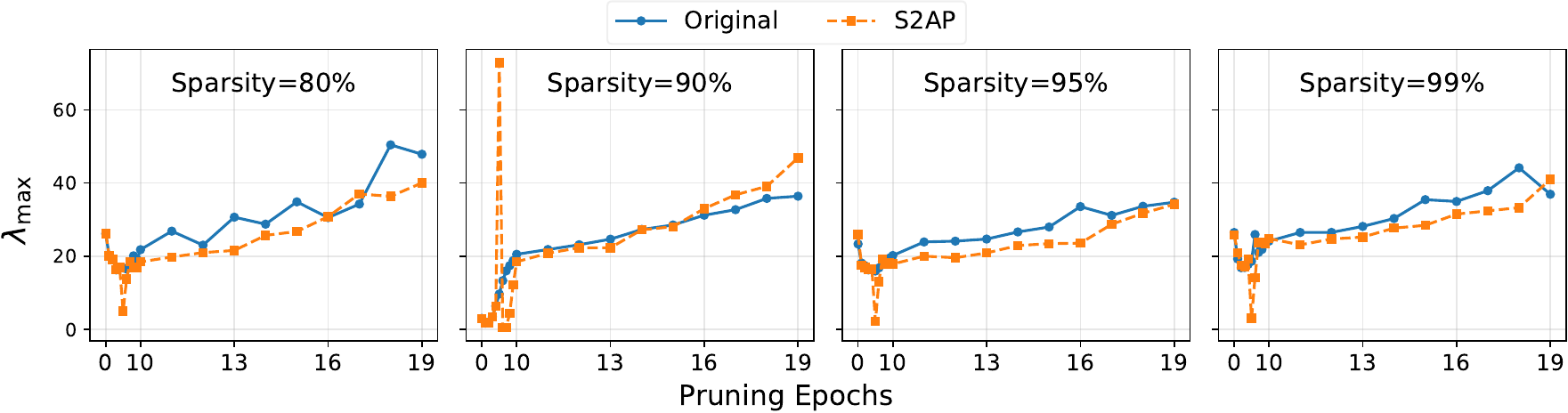}
    \caption{Largest eigenvalue across HARP pruning epochs for ResNet18 on SVHN.}
    \label{fig:apx_eigen_r18_svhn_harp}
\end{figure}

\begin{figure}
    \centering
    \includegraphics[width=1\linewidth]{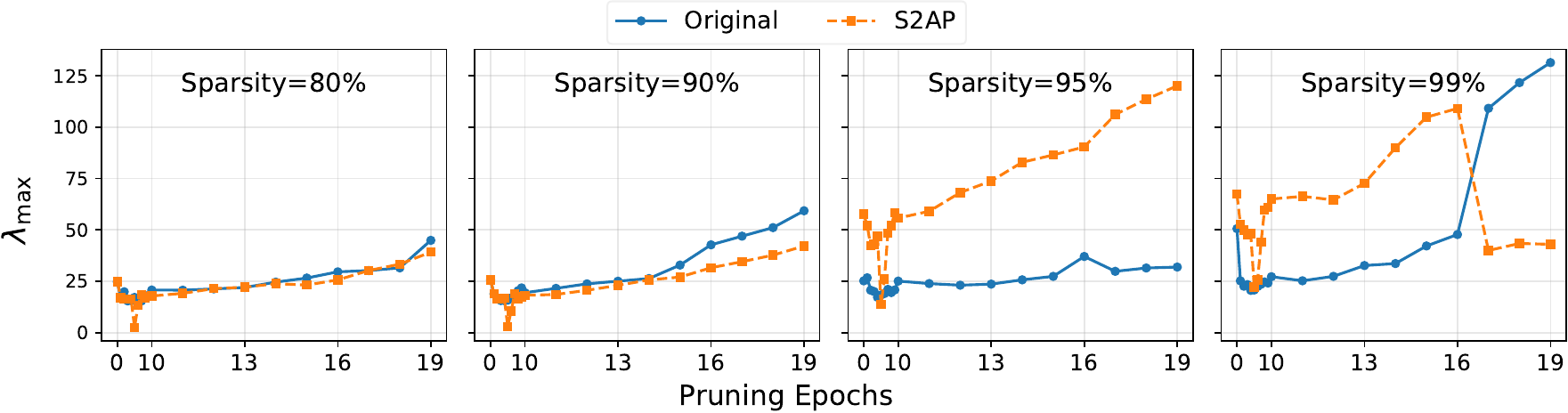}
    \caption{Largest eigenvalue across HYDRA pruning epochs for ResNet18 on SVHN.}
    \label{fig:apx_eigen_r18_svhn_score}
\end{figure}

\begin{figure}
    \centering
    \includegraphics[width=1\linewidth]{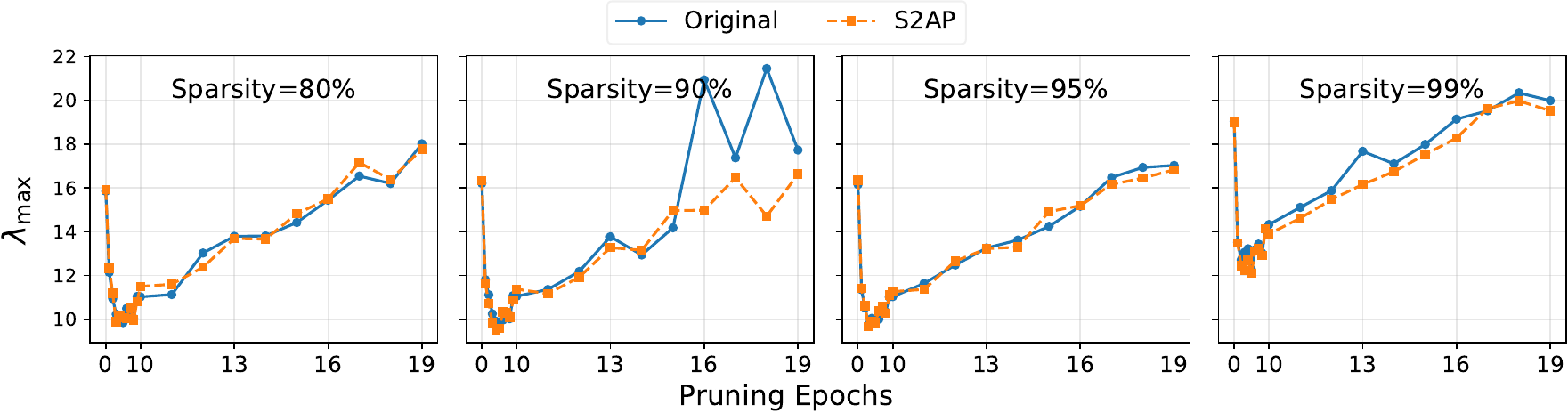}
    \caption{Largest eigenvalue across HARP pruning epochs for VGG on CIFAR10.}
    \label{fig:apx_eigen_vgg16_bn_c10_harp}
\end{figure}

\begin{figure}
    \centering
    \includegraphics[width=1\linewidth]{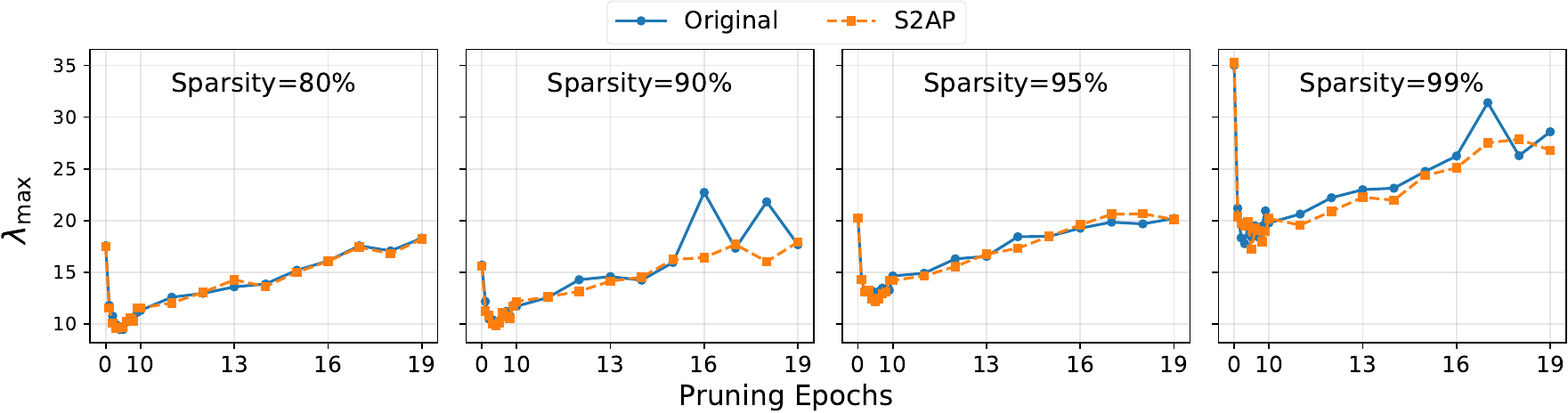}
    \caption{Largest eigenvalue across HYDRA pruning epochs for VGG on CIFAR10.}
    \label{fig:apx_eigen_vgg16_bn_c10_score}
\end{figure}

\begin{figure}
    \centering
    \includegraphics[width=1\linewidth]{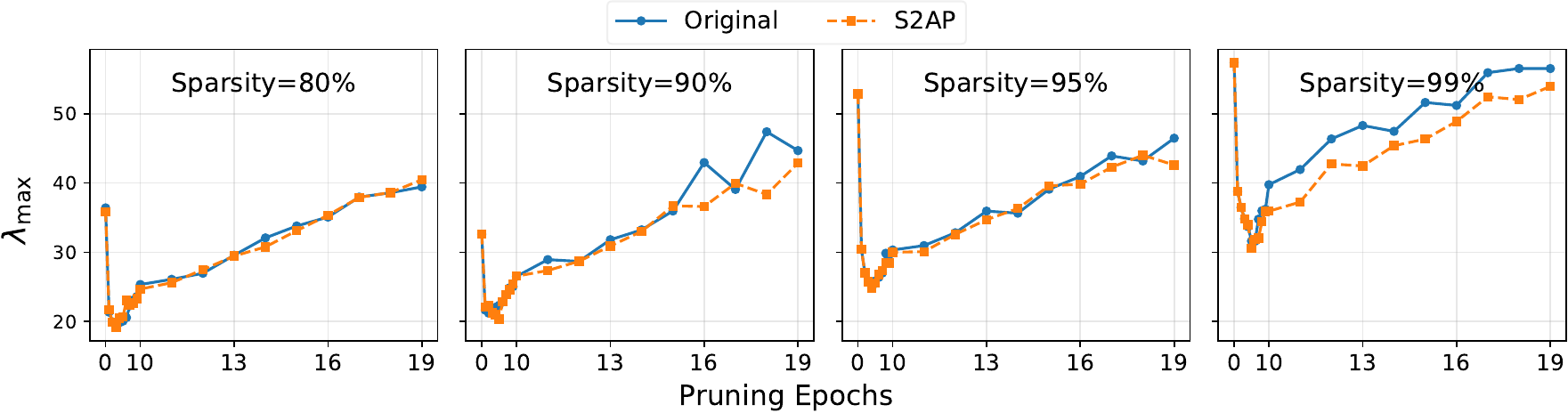}
    \caption{Largest eigenvalue across HARP pruning epochs for VGG on SVHN.}
    \label{fig:apx_eigen_vgg16_bn_svhn_harp}
\end{figure}

\begin{figure}
    \centering
    \includegraphics[width=1\linewidth]{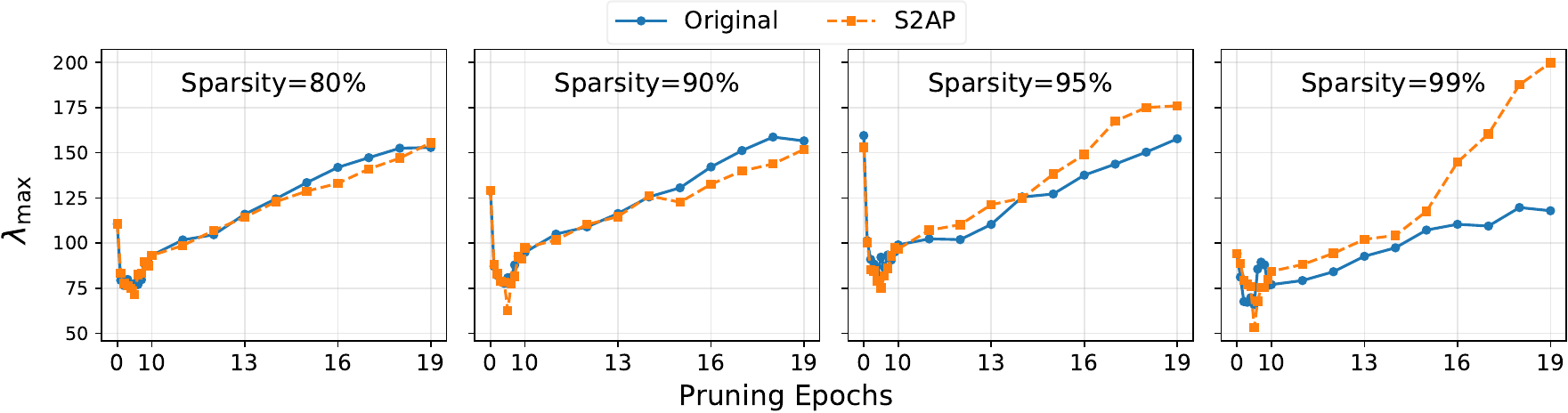}
    \caption{Largest eigenvalue across HYDRA pruning epochs for VGG16 on SVHN.}
    \label{fig:apx_eigen_vgg16_bn_svhn_score}
\end{figure}

\begin{figure}
    \centering
    \includegraphics[width=1\linewidth]{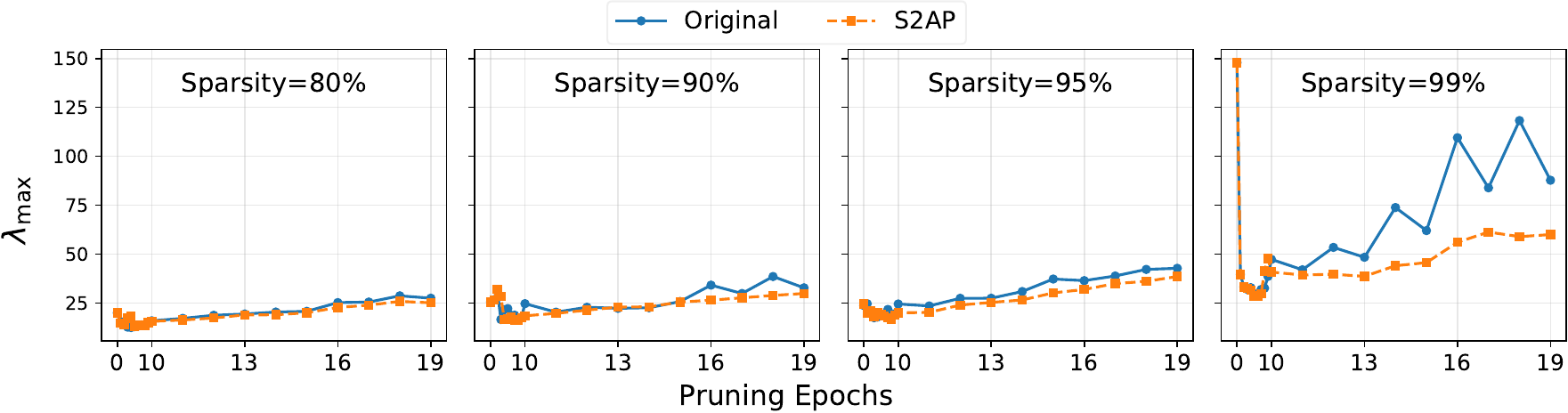}
    \caption{Largest eigenvalue across HYDRA pruning epochs for WideResNet28-4 on CIFAR10.}
    \label{fig:apx_eigen_wrn_c10_score}
\end{figure}

\begin{figure}
    \centering
    \includegraphics[width=1\linewidth]{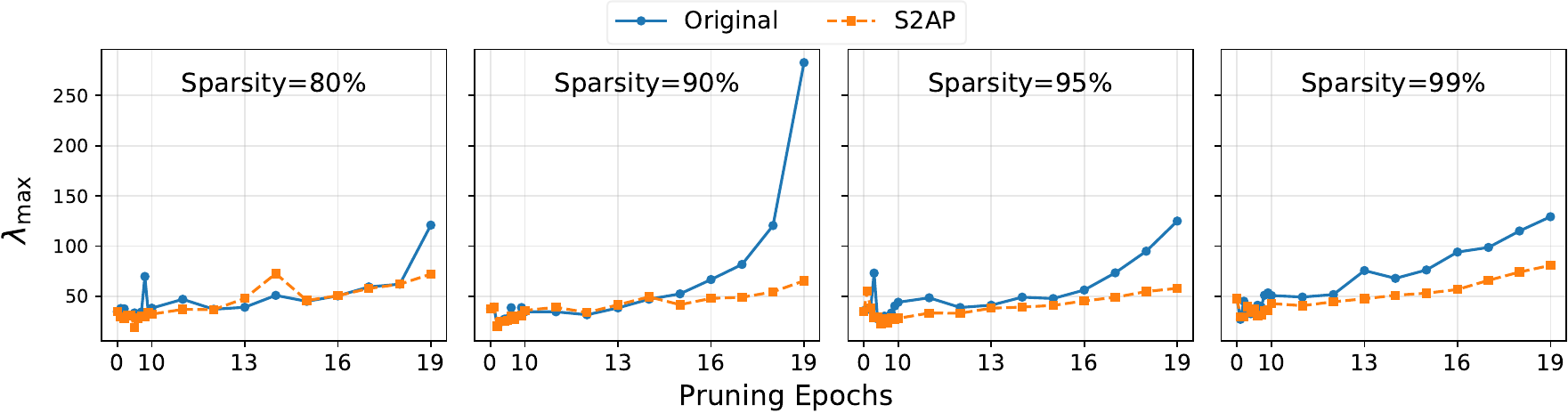}
    \caption{Largest eigenvalue across HARP pruning epochs for WideResNet28-4 on SVHN.}
    \label{fig:apx_eigen_wrn_svhn_harp}
\end{figure}

\begin{figure}
    \centering
    \includegraphics[width=1\linewidth]{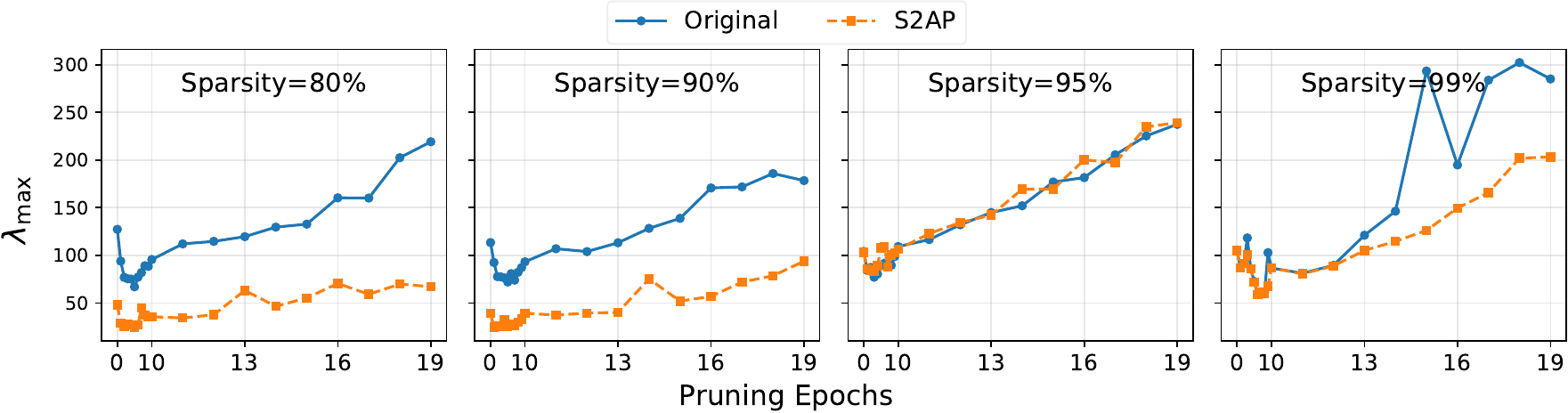}
    \caption{Largest eigenvalue across HYDRA pruning epochs for WideResNet28-4 on SVHN.}
    \label{fig:apx_eigen_wrn_svhn_score}
\end{figure}

To compute the largest eigenvalue of the Hessian $\nabla^2_{\vct s} \mathcal{L}_r(\vct s)$ with respect to the score parameters, we adopt the classical power iteration method. 
Starting from a random unit-norm vector $\vct{v}^{(0)} \in \mathbb{R}^p$, we iteratively compute:
\begin{equation}\label{eq:apx_powerit}
\vct{v}^{(t+1)} = \frac{\nabla^2_s \mathcal{L}_r(\vct{s}) \vct{v}^{(t)}}{\|\nabla^2_s \mathcal{L}_r(\vct{s}) \vct{v}^{(t)}\|_2} \, ,
\end{equation}
where $\mathcal{L}_r(\vct s)=\mathcal{L}_r(\vct w \odot M(\vct s,k), \mathcal{D})$ is the robust loss, that we denote as $\mathcal{L}_r(\vct s)$ to lighten notation.
After $T$ iterations, we compute the Rayleigh quotient as an approximation of the largest eigenvalue:
\begin{equation}\label{eq:apx_lambdamax}
\lambda_{\text{max}} \approx \left\langle \vct{v}^{(T)}, \nabla^2_s \mathcal{L}_r(\vct{s}) \vct{v}^{(T)} \right\rangle \, .
\end{equation}
We select $T=10$ iterations to compute the quotient, and specify that we implement this computation using Hessian-vector products via automatic differentiation, thus refraining from explicitly forming the Hessian~\cite{jastrzkebski2017three}.
This procedure is run at each pruning iteration of both the \ssap and original methods.
We then average the resulting $\lambda_{max}$ values across each iteration and plot the corresponding sharpness trends against epochs.
While we show the CIFAR10 HARP method for ResNet18 in~\autoref{fig:intro_eigenvalues} and for WideResNet28-4 in~\autoref{fig:eigenvalues_wrn}, we complete the remaining plots from~\autoref{fig:apx_eigen_r18_c10_score} to~\autoref{fig:apx_eigen_wrn_svhn_score}.
Overall, the plots show how methods pruned with \ssap hold, apart from a few exceptions, a consistently lower maximum eigenvalue across multiple architectures, datasets, pruning methods, and sparsities. 
We specify how, on the first few epochs, the resulting $\lambda_{max}$ has a negligible difference between Orig. and \ssap methods (hence the first $10$ warped epochs).

\subsection{Measuring Score-Space Loss Difference}\label{app:apdx_lossdiff} 
Measuring sharpness through a loss difference requires perturbing a ``reference'' loss value $\mathcal{L}_r(\vct w \odot M(\vct s,k))$, representing a local minima, through a perturbation $\vct \nu$ which enables measuring sharpness as follows: 
\begin{equation}
    \max_{ \|\vct \nu \odot \vct c^{-1}\|_\infty \leq \rho} \mathcal{L}_r(\vct w \odot M(\vct s + \nu,k), \mathcal{D}) - \mathcal{L}_r(\vct w \odot M(\vct s,k), \mathcal{D}) 
\end{equation}
where $\vct c$ is a positive scaling vector used to make the sharpness definition reparameterization-invariant, addressing the well-known problems of sharpness measures~\cite{dinh17_pmlr}, and the operator $\odot/^{-1}$ defines element-wise multiplication/inversion.
We specify that such a formulation corresponds to the one presented in~\cite{andriushchenko23_icml}, yet adapted to our score-space case.
Overall, we thus perturb the score-space and measure the corresponding loss variation imposed by the shift and mask variation, which we expect to be lower in the \ssap case.

\begin{table}[ht]
\centering
\caption{CIFAR10 Sharpness comparison across sparsity levels and $\rho$ values using Orig. and \ssap pruning strategies. Lower sharpness values are in \textbf{bold}.}
\label{tab:apx_cifar10_sharpness}
\begin{adjustbox}{max width=\textwidth}
\setlength{\tabcolsep}{12pt} 
\renewcommand{\arraystretch}{0.8} 
\begin{tabular}{c ccccccc}
\toprule
\multirow{2}{*}{Model} & Sparsity & $\rho$ & & \multicolumn{2}{c}{\textbf{HARP}} & \multicolumn{2}{c}{\textbf{HYDRA}} \\
& (\%) & & & Orig. & \ssap & Orig. & \ssap \\
\midrule
\multirow{20}{*}{ResNet18} & \multirow{5}{*}{80} 
  & 0.001  & & 0.08316 & \textbf{0.07723} & \textbf{0.08820} & 0.09274 \\
  &        & 0.0025 & & 0.10498 & \textbf{0.09742} & \textbf{0.11074} & 0.11315 \\
  &        & 0.005  & & 0.14170 & \textbf{0.13142} & 0.14845 & \textbf{0.14702} \\
  &        & 0.0075 & & 0.18016 & \textbf{0.16670} & 0.18784 & \textbf{0.18171} \\
  &        & 0.01   & & 0.22096 & \textbf{0.20322} & 0.22839 & \textbf{0.21794} \\
\cmidrule{2-8}
  & \multirow{5}{*}{90} 
  & 0.001  & & 0.07001 & \textbf{0.06848} & 0.08675 & \textbf{0.07637} \\
  &        & 0.0025 & & 0.08566 & \textbf{0.08329} & 0.10258 & \textbf{0.09097} \\
  &        & 0.005  & & 0.11239 & \textbf{0.10844} & 0.13116 & \textbf{0.11596} \\
  &        & 0.0075 & & 0.14069 & \textbf{0.13311} & 0.15879 & \textbf{0.14189} \\
  &        & 0.01   & & 0.16937 & \textbf{0.15981} & 0.18928 & \textbf{0.16741} \\
\cmidrule{2-8}
  & \multirow{5}{*}{95} 
  & 0.001  & & 0.06409 & \textbf{0.06346} & 0.07383 & \textbf{0.06957} \\
  &        & 0.0025 & & 0.07676 & \textbf{0.07504} & 0.08597 & \textbf{0.08035} \\
  &        & 0.005  & & 0.09787 & \textbf{0.09401} & 0.10601 & \textbf{0.09822} \\
  &        & 0.0075 & & 0.11952 & \textbf{0.11332} & 0.12648 & \textbf{0.11706} \\
  &        & 0.01   & & 0.14170 & \textbf{0.13420} & 0.14774 & \textbf{0.13521} \\
\cmidrule{2-8}
  & \multirow{5}{*}{99} 
  & 0.001  & & \textbf{0.05921} & 0.06428 & 0.05573 & \textbf{0.05148} \\
  &        & 0.0025 & & \textbf{0.06637} & 0.07114 & 0.06233 & \textbf{0.05728} \\
  &        & 0.005  & & \textbf{0.07810} & 0.08258 & 0.07365 & \textbf{0.06711} \\
  &        & 0.0075 & & \textbf{0.08930} & 0.09392 & 0.08483 & \textbf{0.07718} \\
  &        & 0.01   & & 0.10852 & \textbf{0.10082} & 0.09631 & \textbf{0.08761} \\
\midrule
\multirow{20}{*}{VGG16} & \multirow{5}{*}{80} 
  & 0.001  & & 0.05925 & \textbf{0.05837} & \textbf{0.05547} & 0.05604 \\
  &        & 0.0025 & & 0.07916 & \textbf{0.07852} & \textbf{0.07089} & 0.07134 \\
  &        & 0.005  & & 0.11328 & \textbf{0.11260} & \textbf{0.09726} & 0.09804 \\
  &        & 0.0075 & & 0.14856 & \textbf{0.14736} & \textbf{0.12450} & 0.12492 \\
  &        & 0.01   & & 0.18579 & \textbf{0.18412} & 0.15311 & \textbf{0.15298} \\
\cmidrule{2-8}
  & \multirow{5}{*}{90} 
  & 0.001  & & \textbf{0.05429} & 0.05503 & 0.05616 & \textbf{0.05531} \\
  &        & 0.0025 & & \textbf{0.07059} & 0.07160 & 0.06881 & \textbf{0.06749} \\
  &        & 0.005  & & \textbf{0.09846} & 0.09970 & 0.08966 & \textbf{0.08804} \\
  &        & 0.0075 & & \textbf{0.12649} & 0.12834 & 0.11178 & \textbf{0.10993} \\
  &        & 0.01   & & \textbf{0.15518} & 0.15657 & 0.13426 & \textbf{0.13282} \\
\cmidrule{2-8}
  & \multirow{5}{*}{95} 
  & 0.001  & & 0.05003 & \textbf{0.04989} & 0.05386 & \textbf{0.04861} \\
  &        & 0.0025 & & 0.06286 & \textbf{0.06237} & 0.06358 & \textbf{0.05832} \\
  &        & 0.005  & & 0.08452 & \textbf{0.08361} & 0.08018 & \textbf{0.07514} \\
  &        & 0.0075 & & 0.10634 & \textbf{0.10562} & 0.09726 & \textbf{0.09266} \\
  &        & 0.01   & & 0.12874 & \textbf{0.12760} & 0.11470 & \textbf{0.11020} \\
\cmidrule{2-8}
  & \multirow{5}{*}{99} 
  & 0.001  & & 0.04367 & \textbf{0.04174} & 0.04815 & \textbf{0.04601} \\
  &        & 0.0025 & & 0.05087 & \textbf{0.04909} & 0.05509 & \textbf{0.05261} \\
  &        & 0.005  & & 0.06309 & \textbf{0.06165} & 0.06656 & \textbf{0.06362} \\
  &        & 0.0075 & & 0.07565 & \textbf{0.07486} & 0.07853 & \textbf{0.07477} \\
  &        & 0.01   & & 0.08851 & \textbf{0.08857} & 0.09058 & \textbf{0.08608} \\
\midrule
\multirow{20}{*}{WRN} & \multirow{5}{*}{80} 
  & 0.001  & & 0.07913 & \textbf{0.07880} & 0.07991 & \textbf{0.07489} \\
  &        & 0.0025 & & 0.09723 & \textbf{0.09593} & 0.09703 & \textbf{0.09096} \\
  &        & 0.005  & & 0.12753 & \textbf{0.12433} & 0.12677 & \textbf{0.11843} \\
  &        & 0.0075 & & 0.15926 & \textbf{0.15295} & 0.15632 & \textbf{0.14677} \\
  &        & 0.01   & & 0.19145 & \textbf{0.18250} & 0.18822 & \textbf{0.17620} \\
\cmidrule{2-8}
  & \multirow{5}{*}{90} 
  & 0.001  & & 0.07006 & \textbf{0.06571} & 0.07159 & \textbf{0.07602} \\
  &        & 0.0025 & & 0.08337 & \textbf{0.07786} & 0.08506 & \textbf{0.08826} \\
  &        & 0.005  & & 0.10571 & \textbf{0.09843} & 0.10735 & \textbf{0.10935} \\
  &        & 0.0075 & & 0.12777 & \textbf{0.11919} & 0.13010 & \textbf{0.13164} \\
  &        & 0.01   & & 0.15019 & \textbf{0.13998} & 0.15363 & \textbf{0.15292} \\
\cmidrule{2-8}
  & \multirow{5}{*}{95} 
  & 0.001  & & \textbf{0.05809} & 0.06162 & 0.06952 & \textbf{0.06229} \\
  &        & 0.0025 & & \textbf{0.06847} & 0.07097 & 0.07998 & \textbf{0.07141} \\
  &        & 0.005  & & \textbf{0.08537} & 0.08655 & 0.09721 & \textbf{0.08766} \\
  &        & 0.0075 & & \textbf{0.10154} & 0.10193 & 0.11500 & \textbf{0.10338} \\
  &        & 0.01   & & 0.11772 & \textbf{0.11709} & 0.13249 & \textbf{0.11976} \\
\cmidrule{2-8}
  & \multirow{5}{*}{99} 
  & 0.001  & & \textbf{0.05823} & 0.05911 & 0.04382 & 0.04925 \\
  &        & 0.0025 & & \textbf{0.06376} & 0.06482 & 0.05483 & \textbf{0.04949} \\
  &        & 0.005  & & \textbf{0.07283} & 0.07452 & 0.06425 & \textbf{0.05900} \\
  &        & 0.0075 & & \textbf{0.08122} & 0.08487 & 0.07172 & \textbf{0.06888} \\
  &        & 0.01   & & 0.08968 & \textbf{0.09446} & 0.08512 & \textbf{0.07535} \\
\bottomrule
\end{tabular}
\end{adjustbox}
\end{table}

In our experiments, we evaluate different $\rho$ values, and show in~\autoref{tab:apx_cifar10_sharpness} an overview of the CIFAR10 results. 
Overall, we see how \ssap consistently reduces sharpness, except for some specific cases at high sparsities. 
In this regard, however, increasing the corresponding $\rho$ value appears to still favor \ssap, suggesting that lower values might not be enough (hence, we choose $\rho=0.01$ in the plot of~\autoref{fig:sharpness_all}).

\subsection{Mask Stability}\label{app:apdx_stability} 
We measure mask stability based on the Hamming distance $\vct h$, which equals measuring the rate of change between masks as follows:
\begin{equation}
    h = \|\vct m_0 \oplus \vct m_t\|_1/|\vct m_0| ,\; \text{where} \;t \in \{1,2, \dots,T\},
\end{equation}
where $\vct m_t$ represents the mask found at epoch $t$, $\oplus$ is the XOR operator measuring the number of differing bits, and $T$ is the total number of epochs. 
We compute $\vct h =\{h_1, h_2, \dots, h_T\}$, thus measuring the distance from the first mask in each epoch, for both original (Orig.) and \ssap adversarial pruning methods. 
Overall, lower $h$ values indicate improved stability, as the number of changed selected weights is, in turn, lower. 
To provide a useful analysis, we compute two vectors, $\vct h_{orig}$ and $\vct h_{\ssap}$, by saving the masks at each epoch while pruning, that we then subtract as $\vct h_{orig}-\vct h_{\ssap}$.  
Hence, we obtain a single curve plot that, when positive, indicates that the \ssap method is more stable than the original one, and vice versa when negative. 

\begin{figure}[htbp]
    \centering
    \begin{tabular}{cc}
        \begin{subfigure}[t]{0.40\textwidth}
            \includegraphics[width=\linewidth]{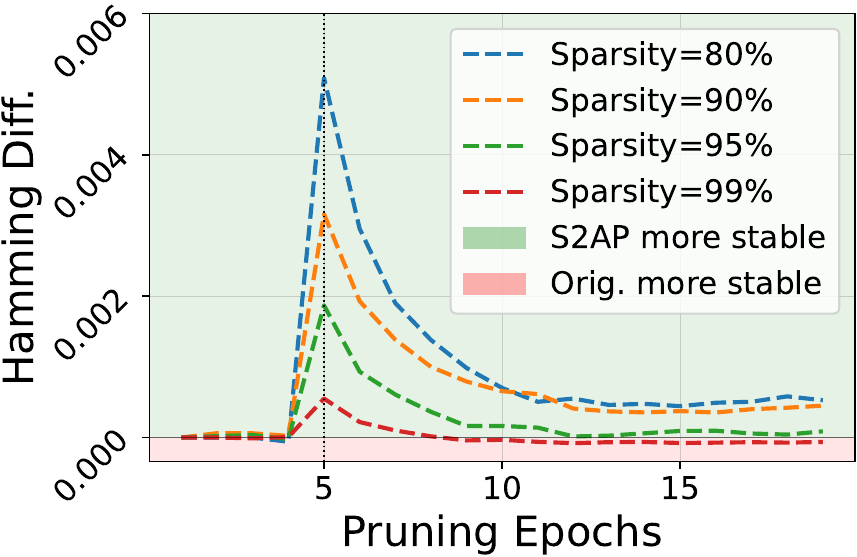}
            \caption{}
        \end{subfigure} &
        \begin{subfigure}[t]{0.40\textwidth}
            \includegraphics[width=\linewidth]{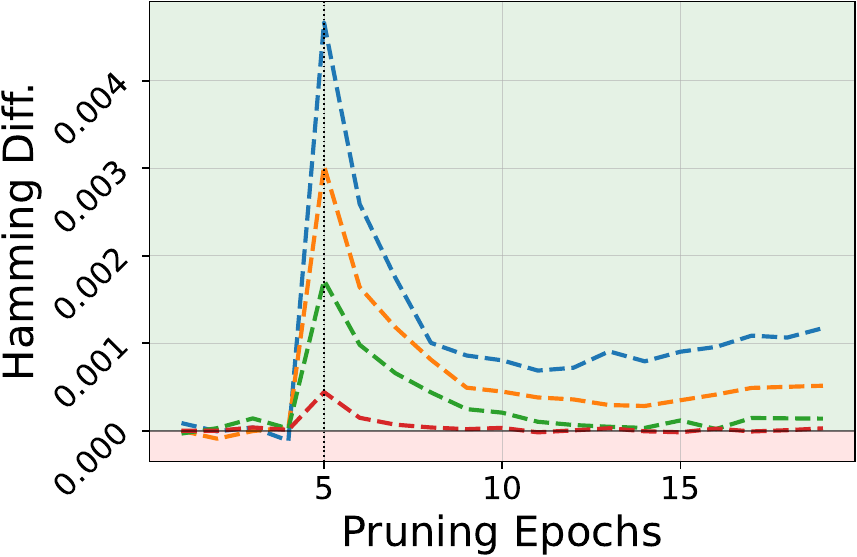}
            \caption{}
        \end{subfigure}
    \end{tabular}
    \caption{Improved mask stability of Resnet18 (a) and WideResNet28-4 (b) on the HYDRA method.}
    \label{fig:apx_stability}
\end{figure}

\begin{figure}[htbp]
    \centering
    \begin{tabular}{cc}
        \begin{subfigure}[t]{0.35\textwidth}
            \includegraphics[width=\linewidth]{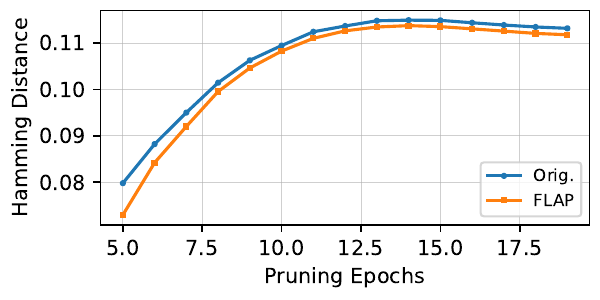}
            \caption{}
        \end{subfigure} &
        \begin{subfigure}[t]{0.35\textwidth}
            \includegraphics[width=\linewidth]{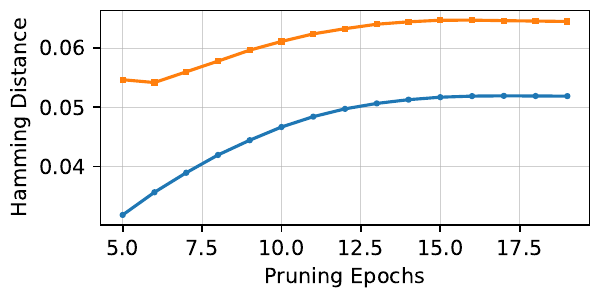}
            \caption{}
        \end{subfigure} \\\\[0.1em]

        \begin{subfigure}[t]{0.35\textwidth}
            \includegraphics[width=\linewidth]{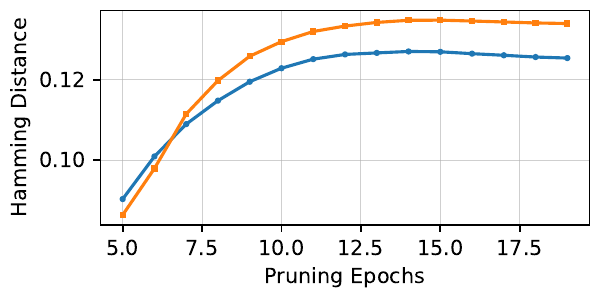}
            \caption{}
        \end{subfigure} &
        \begin{subfigure}[t]{0.35\textwidth}
            \includegraphics[width=\linewidth]{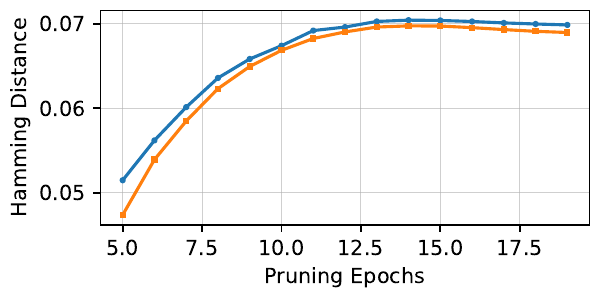}
            \caption{}
        \end{subfigure} \\\\[0.1em]

        \begin{subfigure}[t]{0.35\textwidth}
            \includegraphics[width=\linewidth]{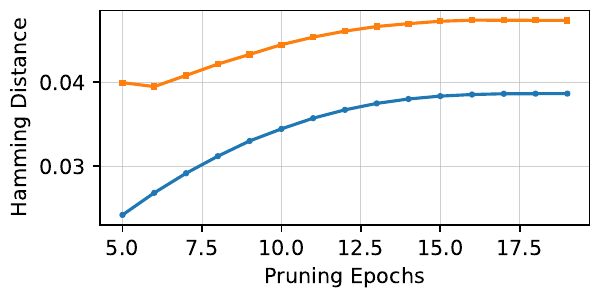}
            \caption{}
        \end{subfigure} &
        \begin{subfigure}[t]{0.35\textwidth}
            \includegraphics[width=\linewidth]{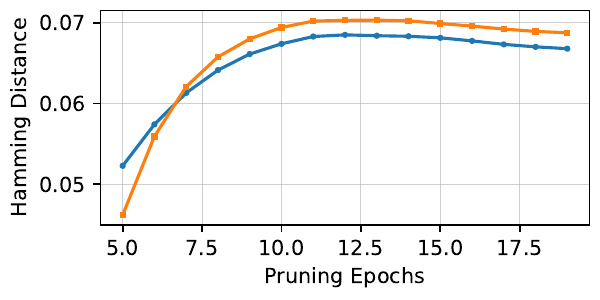}
            \caption{}
        \end{subfigure} \\\\[0.1em]

        \begin{subfigure}[t]{0.35\textwidth}
            \includegraphics[width=\linewidth]{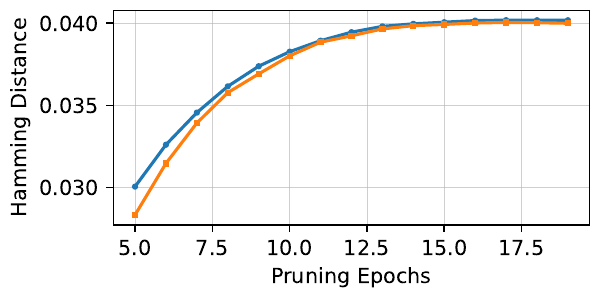}
            \caption{}
        \end{subfigure} &
        \begin{subfigure}[t]{0.35\textwidth}
            \includegraphics[width=\linewidth]{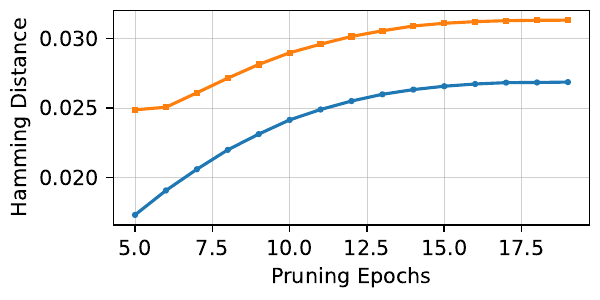}
            \caption{}
        \end{subfigure} \\\\[0.1em]

        \begin{subfigure}[t]{0.35\textwidth}
            \includegraphics[width=\linewidth]{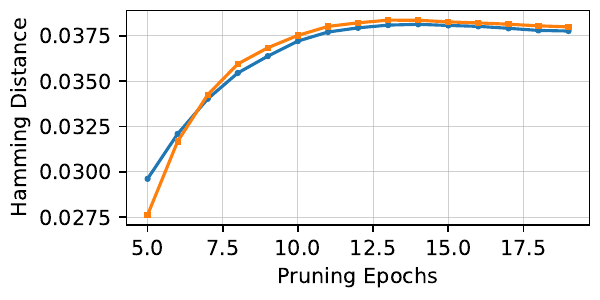}
            \caption{}
        \end{subfigure} &
        \begin{subfigure}[t]{0.35\textwidth}
            \includegraphics[width=\linewidth]{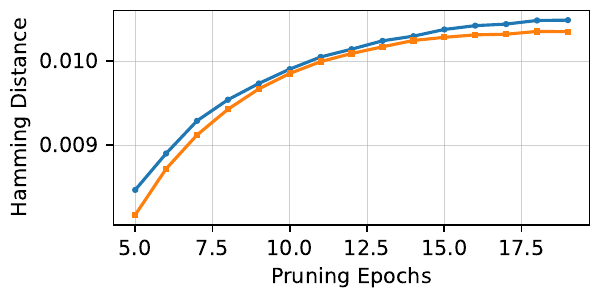}
            \caption{}
        \end{subfigure} \\\\[0.1em]

        \begin{subfigure}[t]{0.35\textwidth}
            \includegraphics[width=\linewidth]{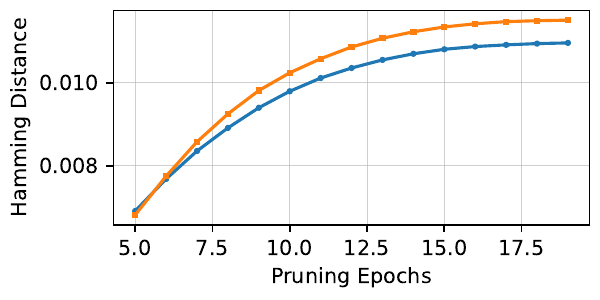}
            \caption{}
        \end{subfigure} &
        \begin{subfigure}[t]{0.35\textwidth}
            \includegraphics[width=\linewidth]{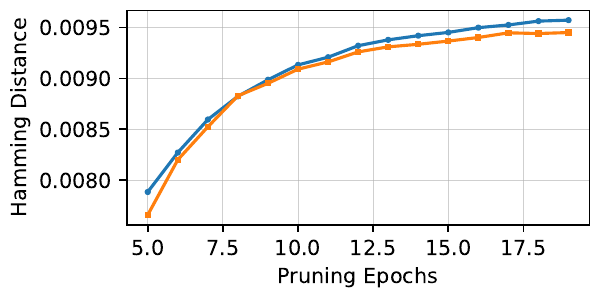}
            \caption{}
        \end{subfigure}
    \end{tabular}
    \caption{Single Hamming distances of VGG16 on CIFAR10 and SVHN after the first $5$ pruning epochs. In (a), (b), and (c) the $80\%$ sparsity for HARP on CIFAR10, HYDRA on CIFAR10, and HARP on SVHN; in (d), (e), and (f) the $90\%$ sparsity for HARP on CIFAR10, HYDRA on CIFAR10, and HARP on SVHN; in (g), (h), and (i) the $95\%$ sparsity for HARP on CIFAR10, HYDRA on CIFAR10, and HARP on SVHN; and in (j), (k), and (l) the $99\%$ sparsity for HARP on CIFAR10, HYDRA on CIFAR10, and HARP on SVHN.}
    \label{fig:apx_vgg_stability}
\end{figure}

Stability is depicted for CIFAR10 HARP method and ResNet18 in~\autoref{fig:intro_hamming}, and for WideResNet28-4 in~\autoref{fig:wrn_mask_stability}. Nonetheless, we provide additional plots for the remaining combinations in~\autoref{fig:apx_stability}, where we show the improved mask stability of \ssap on the HYDRA method as well.
For VGG16 models, interestingly, we find the stability trend often favors the Orig. models instead of \ssap, particularly at lower sparsity. We analyze such a result through the plots of~\autoref{fig:apx_vgg_stability}.
Overall, such a measure allows assessing how much the pruning decisions evolve over time relative to their starting point.



\end{document}